\documentclass[10pt,twocolumn,letterpaper]{article}

\usepackage{caption}
\usepackage{iccv}
\usepackage{times}
\usepackage{epsfig}
\usepackage{graphicx}
\usepackage{amsmath}
\usepackage{amssymb}

\usepackage{array}
\usepackage{comment}
\usepackage[dvipsnames]{xcolor}
\usepackage{multirow,makecell}
\usepackage{gensymb}
\usepackage{booktabs}
\usepackage{nicefrac}
\usepackage{mathtools}
\usepackage{pifont}
\newcommand{\cmark}{\ding{51}}%
\newcommand{\xmark}{\ding{55}}%
\usepackage[inline]{enumitem}
\usepackage[accsupp]{axessibility}  

\makeatletter
\@namedef{ver@everyshi.sty}{}
\makeatother
\usepackage{tikz}
\usetikzlibrary{decorations.pathreplacing,calligraphy}

\usepackage[pagebackref=true,breaklinks=true,letterpaper=true,colorlinks,bookmarks=false]{hyperref}
\usepackage[capitalize]{cleveref}
\crefname{section}{Sec.}{Secs.}
\Crefname{section}{Section}{Sections}
\Crefname{table}{Table}{Tables}
\crefname{table}{Tab.}{Tabs.}
\newcommand{\paragraphNoSpace}[1]{{\bf #1 \,}}

\iccvfinalcopy 

\def\iccvPaperID{****} 

\begin{document}

\title{End2End Multi-View Feature Matching with Differentiable Pose Optimization}

\author{Barbara Roessle and Matthias Nie\ss{}ner \\
Technical University of Munich\\
}

\twocolumn[{%
\renewcommand\twocolumn[1][]{#1}%
\maketitle
\begin{center}
    \captionsetup{type=figure}
    \includegraphics[width=\textwidth,trim={0.2cm 4.2cm 0.6cm 2.7cm},clip]{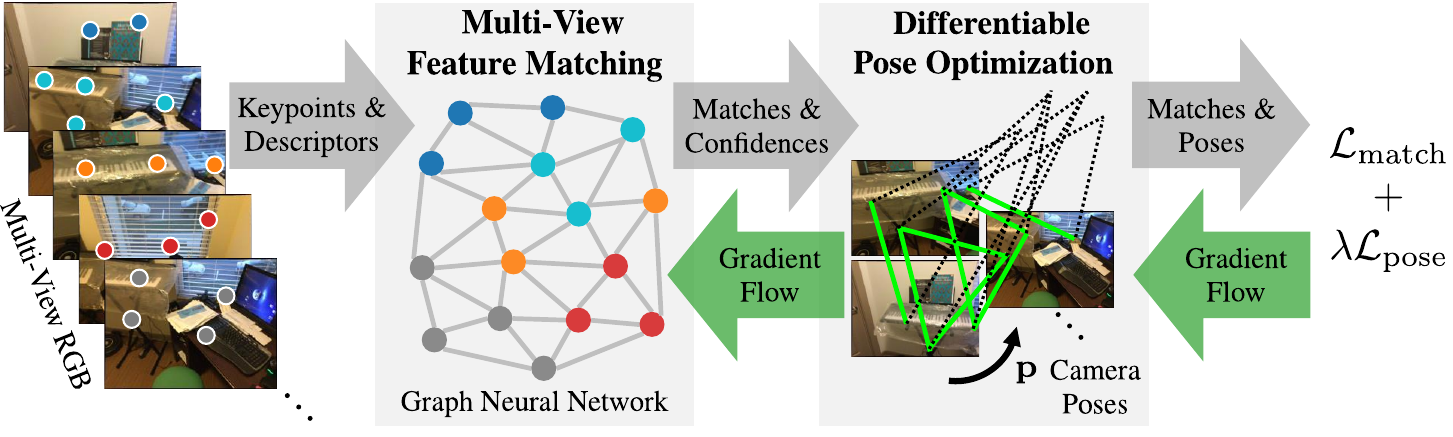}
\end{center}%
\captionof{figure}{We connect feature matching and pose optimization in an end-to-end trainable approach that enables matches and confidence weights to be informed by the pose estimation objective. To this end, we introduce GNN-based multi-view feature matching to predict matches and confidences tailored to a differentiable pose solver, which significantly improves pose estimation performance.}
\label{fig:pipeline}
\vspace{1.0cm}
}]

\ificcvfinal\thispagestyle{empty}\fi

\begin{abstract}
Erroneous feature matches have severe impact on subsequent camera pose estimation and often require additional, time-costly measures, like RANSAC, for outlier rejection. 
Our method tackles this challenge by addressing feature matching and pose optimization jointly.
To this end, we propose a graph attention network to predict image correspondences along with confidence weights. 
The resulting matches serve as weighted constraints in a differentiable pose estimation. Training feature matching with gradients from pose optimization naturally learns to down-weight outliers and boosts pose estimation on image pairs compared to SuperGlue by 6.7\% on ScanNet. At the same time, it reduces the pose estimation time by over 50\% and renders RANSAC iterations unnecessary. 
Moreover, we integrate information from multiple views by spanning the graph across multiple frames to predict the matches all at once. 
Multi-view matching combined with end-to-end training improves the pose estimation metrics on Matterport3D by 18.5\% compared to SuperGlue. 
\end{abstract}


\section{Introduction}
Feature matching is a key component in many 3D vision applications such as structure from motion (SfM) or simultaneous localization and mapping (SLAM). 
Conventional pose estimation is a multi-step process: feature detection finds interest points, for which local descriptors are computed. Based on the descriptors, pairs of keypoints from different images are matched,
which defines constraints in the pose optimization. 
A major challenge lies in the ambiguity of matching local descriptors by nearest-neighbor search, which is error-prone, particularly in texture-less areas or in presence of repetitive patterns.
Hand-crafted heuristics or outlier filters become necessary to circumvent this problem to some degree. 

Recent learning-based approaches~\cite{Sarlin2020SuperGlueLF,Sun2021LoFTRDL,Jiang2021COTRCT,Mao20223DGSTFM}
instead leverage the greater image context to improve the matching, e.g.,
SuperGlue~\cite{Sarlin2020SuperGlueLF} introduces a graph neural network (GNN) for descriptor matching on an image pair. 
Graph edges connect keypoints from arbitrary locations
and enable reasoning in a broad context, leading to globally well-informed solutions compared to convolutional neural networks (CNN) with limited receptive field. 
The receptive field in SuperGlue, however, remains limited by the two-view setup, despite that more images are typically available in pose estimation tasks. 
Our idea is to further facilitate information flow by joining  
multiple views in the matching process. This way, we allow multi-view correlation to strengthen geometric reasoning and confidence prediction. 
Joint matching of multiple images integrates well into pose estimation pipelines, as they typically solve for more than two cameras. 

Additionally, we note that accurate feature matching, in and of itself, does not necessarily give rise to accurate pose estimation, as the spatial distribution of feature matches is essential for robust pose optimization.
For instance, perfectly precise matches may form a degenerate case (e.g., lying on a line) and thus have no value for pose optimization. 
In addition, confidence scores predicted by matching networks do not necessarily reflect the value of matches towards pose optimization. 
Feature matching and pose estimation are thus tightly coupled problems, for which we propose a joint solution. 

We encode keypoints and descriptors from multiple images to construct a graph, where self-attention provides context awareness within the same image and cross-attention enables reasoning with respect to all other images. A GNN predicts matches along with confidence weights, which define constraints on the camera poses that we optimize with a differentiable solver. The GNN is trained end-to-end using gradients from the pose optimization. From this feedback, the network learns to produce valuable matches for pose estimation and thereby 
learns effective outlier rejection. 
We evaluate our method on image pairs and in a multi-view setting on ScanNet~\cite{Dai2017ScanNetR3}, Matterport3D~\cite{Chang2017Matterport3DLF}, and MegaDepth~\cite{Li2018MegaDepthLS} datasets and show that our joint approach to feature matching and pose estimation improves over prior work on learned feature matching, enabled by the following contributions: 
\begin{itemize}
    \item We introduce an end-to-end trainable pose estimation that both guides confidence weights of feature matches in an unsupervised fashion and backpropagates gradients to inform the matching network. 
    \item We propose a multi-view graph attention network to learn feature matches simultaneously across multiple frames. 
\end{itemize}

\section{Related Work}
\paragraphNoSpace{Conventional Feature Matching.}
The classical feature matching pipeline comprises the following steps: 1) interest point detection, 2) feature description, 3) matching through nearest neighbor search in descriptor space, and 4) outlier filtering. In this pipeline, hand-crafted features like SIFT~\cite{LoweDavid2004DistinctiveIF} and ORB~\cite{Rublee2011ORBAE} are very successful and have been widely used for many years. However, they tend to struggle with appearance or viewpoint changes. 
Starting with LIFT~\cite{Yi2016LIFTLI}, learning-based descriptors have been developed to tackle these challenges~\cite{Ono2018LFNetLL,Dusmanu2019D2NetAT,Revaud2019R2D2RA,Bhowmik2020ReinforcedFP,Tyszkiewicz2020DISKLL}. They often combine interest point detection and description, such as SuperPoint \cite{DeTone2018SuperPointSI}, which we use for our method. 
Nearest neighbor feature matching is prone to outliers, making post-processing methods indispensable. This includes mutual check, ratio test \cite{LoweDavid2004DistinctiveIF}, neighborhood consensus \cite{Tuytelaars2000WideBS,Cech2008EfficientSC,Cavalli2020HandcraftedOD,Bian2017GMSGM,Ma2018LocalityPM} and sampling-based outlier rejection~\cite{Fischler1981RandomSC,Barth2019MAGSACMS,Raguram2008ACA}. 
Learning-based approaches have also addressed outlier detection~\cite{Yi2018LearningTF,Ranftl2018DeepFM,Brachmann2019NeuralGuidedRL,Zhang2019LearningTC}---these methods rely on reasonable matching proposals and lack visual information in their decision process. 

\paragraphNoSpace{Learning Feature Matching.}
Recent methods employ neural networks for feature matching on image pairs. There are methods that determine dense, pixel-wise correspondences 
with confidence estimates for filtering~\cite{Rocco2018NeighbourhoodCN,Rocco2020EfficientNC,Li2020DualResolutionCN}. 
However, the matching lacks global context due to the limited receptive field of CNNs and fails to distinguish regions of little texture or repetitive structure. 
In contrast, SuperGlue~\cite{Sarlin2020SuperGlueLF} represents a sparse matching network that operates on keypoints with descriptors. Using an attentional GNN~\cite{Vaswani2017AttentionIA} all keypoints interact, hence the receptive field spans across both images, leading to accurate matches in wide-baseline settings. Inspired by 
GNN-based feature matching, we build upon SuperGlue by enhancing its receptive field through multi-view matching and by improving outlier filtering through end-to-end training with pose optimization. 
LoFTR~\cite{Sun2021LoFTRDL} and COTR~\cite{Jiang2021COTRCT} recently proposed detector-free methods that operate on RGB images directly. Using attention and a coarse-to-fine approach, they equally achieve a receptive field across the image pair and high quality matches. 
3DG-STFM~\cite{Mao20223DGSTFM} extends LoFTR with student-teacher learning to leverage 
RGB-comprised depth information. 
We show that our end-to-end and multi-view approach improves pose estimation over SuperGlue and the detector-free methods LoFTR, COTR, and 3DG-STFM. 

\paragraphNoSpace{Pose Optimization.}
Once matches between a set of images are found, bundle adjustment formulations~\cite{triggs1999bundle} are used to optimize poses on RGB~\cite{agarwal2011building} or RGB-D data~\cite{dai2017bundlefusion}.
This typically leads to non-linear least squares problems which are optimized with non-linear solvers, like Gauss-Newton or Levenberg-Marquardt. 
Such pipelines usually perform feature matching as a pre-process, followed by a filtering with a combination of RANSAC and robust optimization techniques~\cite{zach2014robust,Choi_2015_CVPR}. 
However, feature matching and pose optimization largely remain separate steps and cannot inform each other.
To this end, differentiable pose optimization techniques, such as DeMoN~\cite{ummenhofer2017demon}, BA-Net~\cite{tang2018ba}, RegNet~\cite{han2018regnet}, or 3DRegNet~\cite{pais20203dregnet}, propose to obtain gradients through the pose optimization that in turn guide the  learning of feature descriptors. 
In contrast to treating feature extraction as a separate step, feature descriptors are then learned with the objective to obtain well-aligned poses. 
In our work, we go a step further and focus on learning how to match features rather than using a predefined matching method.
We leverage differentiable pose optimization to provide gradients for our feature matching network, and achieve significantly improved pose estimation results.

\section{Method}
Our method associates keypoints from $N$ images $\{I_n\}^{N}_{n=1}$, such that the resulting matches and confidence weights are particularly valuable for estimating the corresponding camera poses $\{\mathbf{p}_n\}^{N}_{n=1}$; $\mathbf{p}_n \in \mathbb{R}^6$. 
Keypoints are represented by their image coordinates $\mathbf{x} \in \mathbb{R}^2$, visual descriptors $\mathbf{d} \in \mathbb{R}^D$ and a confidence score $c \in [0, 1]$. 
We use the SuperPoint network for feature detection and description \cite{DeTone2018SuperPointSI}. 
Our pipeline (\cref{fig:pipeline}) ties together feature matching and pose optimization: we employ a GNN to associate keypoints across multiple images (\cref{ssec:multi_view_matching}). The resulting matches and confidence weights define constraints in the subsequent pose optimization (\cref{ssec:pose_optimization}), which is differentiable, thus enabling end-to-end training (\cref{ssec:end2end_training}). Both, multi-view and end-to-end, are independent and can be used in isolation, however, the benefit is larger in combination, as shown in the experiments (\cref{sec:results}). 

\subsection{Multi-View Graph Attention Network}
\label{ssec:multi_view_matching}
\paragraphNoSpace{Motivation.}
In the multi-view matching problem of $N$ images, each keypoint matches to at most $N - 1$ other keypoints, where each of the matching keypoints belongs to a different input image. 
Without knowing the transformations between images, one keypoint can match to any keypoint location in the other images. Hence, all keypoints in the other images need to be considered as matching candidates. Although keypoints from the same image are not matching candidates, they contribute valuable constraints in the assignment problem, e.g., their projection into other images must follow consistent transformations. The matching problem can be represented as a graph, where nodes model keypoints and edges their relationships. 
A GNN architecture reflects this structure and enables learning the complex relations between keypoints to determine feature matches. The iterative message passing process enables the search for globally optimal matches as opposed to a greedy local assignment. 
On top of that, attention-based message aggregation allows each keypoint to focus on information from the keypoints that provide the most  insight for its assignment. 
We build upon SuperGlue, which introduces an attentional GNN for descriptor matching on image pairs \cite{Sarlin2020SuperGlueLF}. Our extension to multi-image matching is motivated by the following: first,  graph-based reasoning can benefit from tracks that are longer than two keypoints---i.e., a match becomes more confident, if multiple views agree on the keypoint similarity and its coherent location with respect to the other keypoints in each frame. 
In particular, with regards to robust pose optimization, it is crucial to facilitate this information flow and boost the confidence prediction. Second, pose estimation or SLAM systems generally consider multiple input views. 
With the described graph structure, jointly 
matching $N$ images is more efficient in terms of intra-frame GNN messages than matching the corresponding image  pairs individually, as detailed in the supplementary material. 

\begin{figure}[tbp]
\begin{center}
\includegraphics[width=\linewidth,trim={1.5cm 6.8cm 1.cm 4.cm},clip]{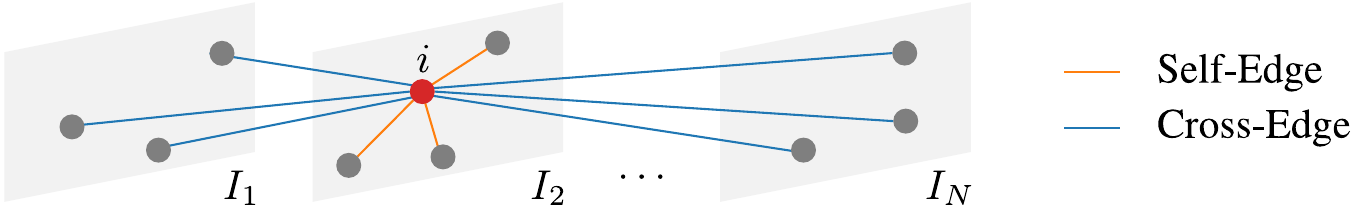}
\end{center}
\caption{Keypoints are graph nodes. Keypoint $i$ is connected to keypoints in the same image through self-edges and to keypoints in other images though cross-edges.}
\label{fig:graph_edges}
\end{figure}

\paragraphNoSpace{Graph Construction.}
Each keypoint represents a graph node. The initial node embedding ${}^{(1)}\mathbf{f}_i$ of keypoint $i$ is computed from its image coordinate $\mathbf{x}_i$, confidence $c_i$ and descriptor $\mathbf{d}_i$, which allows 
the GNN to consider spatial location, certainty and visual appearance in the matching: 
\begin{equation}
    {}^{(1)}\mathbf{f}_i = \mathbf{d}_i + F_\mathrm{encode}\left(\left[\mathbf{x}_i \mathbin\Vert c_i\right]\right),
    \label{eq:node_embedding}
\end{equation}
where $\mathbin\Vert$ denotes row-wise concatenation. $F_{\mathrm{encode}}$ is a multilayer perceptron (MLP) that lifts the image point and its confidence into the high-dimensional space of the descriptor to help 
the spatial learning \cite{Sarlin2020SuperGlueLF,Gehring2017ConvolutionalST,Vaswani2017AttentionIA}. 
The graph nodes are connected by two kinds of edges: self-edges connect keypoints within the same image. Cross-edges connect keypoints from different images (\cref{fig:graph_edges}). The edges are undirected, i.e., information flows in both directions. 

\paragraphNoSpace{Message Passing.}
Interaction between keypoints---the graph nodes---is realized through message passing \cite{Duvenaud2015ConvolutionalNO,Gilmer2017NeuralMP}. The goal is to achieve a state where node descriptors of matching keypoints are close in descriptor space, whereas unrelated keypoints are far apart. The GNN has $L$ layers, where each layer $\ell$ corresponds to a message exchange between keypoints. The layers alternate between updates along self-edges $\mathcal{E}_{\mathrm{self}}$ and cross-edges $\mathcal{E}_{\mathrm{cross}}$---starting with an exchange along self-edges in layer $\ell=1$~\cite{Sarlin2020SuperGlueLF}. \cref{eq:node_update} describes the iterative node descriptor update, where ${}^{(\ell)}\mathbf{m}_{\mathcal{E}\rightarrow i}$ is the aggregated message from all keypoints that are connected to keypoint $i$ by an edge in $\mathcal{E} \in \{\mathcal{E}_{\mathrm{self}}, \mathcal{E}_{\mathrm{cross}}\}$. 
${}^{(\ell)}F_{\mathrm{update}}$ is a MLP, where each GNN layer $\ell$ has a separate set of network weights. 
\begin{equation}
    {}^{(\ell+1)}\mathbf{f}_i = {}^{(\ell)}\mathbf{f}_i + {}^{(\ell)}F_{\mathrm{update}}\left(\left[{}^{(\ell)}\mathbf{f}_i \mathbin\Vert {}^{(\ell)}\mathbf{m}_{\mathcal{E}\rightarrow i}\right]\right)
    \label{eq:node_update}
\end{equation}
Multi-head attention~\cite{Vaswani2017AttentionIA} is used to merge all incoming information for keypoint $i$ into a single message ${}^{(\ell)}\mathbf{m}_{\mathcal{E}\rightarrow i}$~\cite{Sarlin2020SuperGlueLF}. 
Messages along self-edges are combined by self-attention between the keypoints of the same image, messages along cross-edges by cross-attention between the keypoints from all other images. 
Linear projection of node descriptors is used to compute the query ${}^{(\ell)}\mathbf{q}_i$ of query keypoint $i$, as well as the keys ${}^{(\ell)}\mathbf{k}_j$ and values ${}^{(\ell)}\mathbf{v}_j$ of its source keypoints $j$:
\begin{align}
    {}^{(\ell)}\mathbf{q}_i &= {}^{(\ell)}\mathbf{W}_1 {}^{(\ell)}\mathbf{f}_i + {}^{(\ell)}\mathbf{b}_1 ,
    \label{eq:query} \\
    \begin{bmatrix}
    {}^{(\ell)}\mathbf{k}_j \\ {}^{(\ell)}\mathbf{v}_j
    \end{bmatrix}
    &= 
    \begin{bmatrix}
    {}^{(\ell)}\mathbf{W}_2 \\ {}^{(\ell)}\mathbf{W}_3
    \end{bmatrix}
    {}^{(\ell)}\mathbf{f}_j +
    \begin{bmatrix}
    {}^{(\ell)}\mathbf{b}_2 \\ {}^{(\ell)}\mathbf{b}_3
    \end{bmatrix}.
    \label{eq:key_val}
\end{align}
The set of source keypoints $\{j : (i, j) \in \mathcal{E}\}$ comprises all keypoints connected to $i$ by an edge of the type, that is relevant to the current layer. $\mathbf{W}$ and $\mathbf{b}$ are per-layer weight matrices and bias vectors, respectively.
For each source keypoint the similarity to the query is computed by the dot product ${}^{(\ell)}\mathbf{q}_i\cdot{}^{(\ell)}\mathbf{k}_j$. The softmax over the similarity scores determines the attention weight $\alpha_{ij}$ of each source keypoint $j$ in the aggregated message to $i$: 
\begin{equation}
    {}^{(\ell)}\mathbf{m}_{\mathcal{E}\rightarrow i} = \sum_{j : (i, j) \in \mathcal{E}} {}^{(\ell)}\alpha_{ij}{}^{(\ell)}\mathbf{v}_j .
\end{equation}
It is important to note that in cross-attention layers, 
the source keypoints $j$ to a query keypoint $i$ come from multiple images. The softmax-based weighting is robust to variable number of input views and therewith variable number of keypoints. 
After $L$ message passing iterations the node descriptors for subsequent assignment are retrieved by linear projection: 
\begin{equation}
    \mathbf{f}_i = \mathbf{W}_4 {}^{(L+1)}\mathbf{f}_i + \mathbf{b}_4 .
    \label{eq:final_proj}
\end{equation}

\paragraphNoSpace{Partial Assignment.}
The partial assignment problem of keypoints from two images can be solved with the differentiable Sinkhorn algorithm \cite{Sinkhorn1967ConcerningNM,Cuturi2013SinkhornDL,Sarlin2020SuperGlueLF}: Given an input score matrix, a partial assignment is optimized, where each keypoint either obtains a match in the other image or remains unmatched. 
We compute the assignment on the set of possible image pairs $\mathcal{P}$, 
excluding pairs between identical images and pairs that are a permutation of another pair. For each pair $(a,b) \in \mathcal{P}; a,b \in \{1,2,\dots,N\}$, the score matrix is filled with the dot-product similarities of node descriptors. 
From the resulting partial assignment matrix $\mathbf{P}_{ab}$, the set of matches is derived: first, a candidate match for each keypoint is determined by the row-wise and column-wise maximal elements. 
Second, we keep only those matches, where both keypoints mutually agree on the assignment.

\paragraphNoSpace{Confidence Prediction.}
For each pair of matching keypoints $i,j$ a confidence weight $w_{ij}$ is predicted from the final node descriptors $\mathbf{f}_i,\mathbf{f}_j$ and their score in the corresponding partial assignment matrix $\mathbf{P}_{ab}$: 
\begin{equation}
    w_{ij} = F_\mathrm{conf\_1}(F_\mathrm{conf\_2}(\mathbf{P}_{ab,i,j}) + F_\mathrm{conf\_3}\left(\left[\mathbf{f}_i \mathbin\Vert \mathbf{f}_j\right]\right)),
    \label{eq:conf_mlp}
\end{equation}
where $F_\mathrm{conf\_*}$ represent small MLPs. 
\subsection{Differentiable Pose Optimization}
We introduce a differentiable relative pose optimization that provides supervision signal for feature matching. 
It is composed of two parts: initial pose estimation through a weighted eight-point algorithm and pose refinement through bundle adjustment. 
\label{ssec:pose_optimization}

\paragraphNoSpace{Weighted Eight-Point Algorithm.}
For each image pair, a fundamental matrix $\mathbf{F}$ 
is computed using the eight-point algorithm~\cite{LonguetHiggins1981ACA} with input coordinate normalization~\cite{Hartley1997InDefense}. 
To facilitate the learning of meaningful confidences, it is essential to consider all matches in a weighted manner. Hence, we define the system of linear equations as a confidence-weighted version of the eight-point algorithm: 
\begin{equation}
    \mathrm{diag}(\mathbf{w})\mathbf{A}\,\mathrm{flat}(\mathbf{F})= \mathbf{0}.
    \label{eq:8pt_linear_system}
\end{equation}
\cref{eq:8pt_linear_system} follows from the epipolar geometry $\mathbf{x}'^{\top} \mathbf{F} \mathbf{x}=0$ by arranging the known coordinates of a match, $\mathbf{x}=[x,y,z]^{\top}$ and $\mathbf{x'}=[x',y',z']^{\top}$, into matrix $\mathbf{A}$ and flattening $\mathbf{F}$ in column-major order to a vector $\mathrm{flat}(\mathbf{F})$. 
Each row $[xx',xy',x,yx',yy',y,x',y',1]$ in $\mathbf{A}$ describes one match and is multiplied with its confidence through the diagonal matrix $\mathrm{diag}(\mathbf{w})$ from the vector of confidences $\mathbf{w}$. 
Given more than 8 matches, the system is overdetermined. Thus, we search a least-squares solution for $\mathbf{F}$ that minimizes $\Vert \mathrm{diag}(\mathbf{w})\mathbf{A}\,\mathrm{flat}(\mathbf{F}) \Vert_2$ under the constraint $\Vert \mathrm{flat}(\mathbf{F}) \Vert_2=1$ to avoid the trivial solution. 
Singular value decomposition (SVD) of $\mathrm{diag}(\mathbf{w})\mathbf{A}$ determines this solution as the singular vector with the smallest singular value and we force the resulting $\mathbf{F}$ to have rank 2~\cite{hartley_zisserman_2004}. The partial derivatives of the SVD can be computed in closed-form~\cite{IonescuMatrixBackpropagation}, thus the eight-point algorithm suits well for end-to-end training. 
Given the intrinsics and the resulting $\mathbf{F}$, there are four possible solutions for the relative transformation between an image pair, aside from unknown scale. During training, we select the solution closest to the ground truth. At test time, following the cheirality constraint \cite{hartley_zisserman_2004}, the solution with most triangulated points in front of both cameras is chosen. 

\paragraphNoSpace{Bundle Adjustment.}
The initial relative pose $\mathbf{p}_{\mathrm{init}}$ from the weighted eight-point algorithm is refined using a bundle adjustment formulation. To this end, we introduce a differentiable optimizer $\Omega$ to refine the relative pose $\mathbf{p}$ and estimate 3D points $\mathbf{Y} \in \mathbb{R}^{M\times 3}$ for the matches $\mathcal{M}$:
\begin{equation}
    \{\mathbf{p}, \mathbf{Y}\} = \Omega(\mathbf{p}_{\mathrm{init}},\mathcal{M}).
    \label{eq:optimizer}
\end{equation}
For each match $m$, we compute confidence-weighted residuals $\mathbf{r}_m, \mathbf{r}_m' \in \mathbb{R}^2$ on the projection of the corresponding 3D point $\mathbf{y}$ into each image and define the energy as the sum of squares:
\begin{align}
    E(\mathbf{p},\mathbf{Y})=\!\!\!\!\!\!\sum_{\substack{(\mathbf{x},\mathbf{x}',w), \mathbf{y} \,\in\, \mathcal{M}, \mathbf{Y}}}\!\!\!\!\!\! \left(\left\Vert\mathbf{r}_{m}\right\Vert_2^2 + \left\Vert\mathbf{r}_m'\right\Vert_2^2\right), \enspace \text{where} \\
    \mathbf{r}_m = w\left(\pi(\mathbf{y})-\mathbf{x}\right) , \enspace\mathbf{r}_m' = w\left(\pi'(\mathbf{R}\mathbf{y}+\mathbf{t})-\mathbf{x}'\right).
    \label{eq:energy}
\end{align}
$\mathbf{x}$ and $\mathbf{x}'$ are the image coordinates of a match and $w$ is its confidence. The 3D points are defined in 
the first camera frame and $\{\mathbf{R} \in \mathbb{R}^{3\times3},\mathbf{t} \in \mathbb{R}^{3}\}$ describes the transformation 
from the first to the second camera, for which $\mathbf{p} \in \mathbb{R}^{6}$ is the equivalent pose vector in $\mathfrak{se}(3)$ coordinates, i.e., three translation elements followed by three rotation elements.
The functions, $\pi$ and $\pi'$, project a 3D point from the respective camera frame to its image plane. $\mathbf{p}$ is initialized to $\mathbf{p}_{\mathrm{init}}$ and $\mathbf{Y}$ is initialized by triangulating the matches. 

Gauss-Newton algorithm is used to minimize the energy with respect to the relative pose and the 3D points. Thus, we optimize for a vector $\mathbf{z}=\left[\mathbf{p} \mathbin\Vert \mathrm{flat}(\mathbf{Y}^{\top})\right] \in \mathbb{R}^{6+3M}$ and compose a residual vector $\mathbf{r}=\left[\mathbf{r}_1 \mathbin\Vert \mathbf{r}_1' \mathbin\Vert \dots \mathbin\Vert \mathbf{r}_M \mathbin\Vert \mathbf{r}_M'\right] \in \mathbb{R}^{4M}$, where $M$ is the number of matches. 
The Jacobian matrix $\mathbf{J}\in \mathbb{R}^{4M\times(6+3M)}$ is initialized to $\mathbf{0}$ and for each match $m$ the corresponding submatrices are filled with the partial derivatives with respect to the pose $\frac{\partial \mathbf{r}_m'}{\partial \mathbf{p}} \in \mathbb{R}^{2\times6}$ and with respect to the 3D point $\frac{\partial \mathbf{r}_m}{\partial \mathbf{y}}, \frac{\partial \mathbf{r}_m'}{\partial \mathbf{y}} \in \mathbb{R}^{2\times3}$ \cite{Blanco}: 
\begin{align}
    \frac{\partial \mathbf{r}_m'}{\partial \mathbf{p}}&=w\frac{\partial \pi'(\mathbf{R}\mathbf{y}+\mathbf{t})}{\partial (\mathbf{R}\mathbf{y}+\mathbf{t})}\begin{bmatrix}
        \mathbf{I} & -\left(\mathbf{R}\mathbf{y}+\mathbf{t}\right)^\wedge
    \end{bmatrix}, \label{eq:jac_cam} \\
    \frac{\partial \mathbf{r}_m}{\partial \mathbf{y}}&=w \frac{\partial \pi(\mathbf{y})}{\partial \mathbf{y}} \;,\quad\quad
    \frac{\partial \mathbf{r}_m'}{\partial \mathbf{y}}=w \frac{\partial \pi'(\mathbf{R}\mathbf{y}+\mathbf{t})}{\partial (\mathbf{R}\mathbf{y}+\mathbf{t})} \mathbf{R}\;,
    \label{eq:jac_pts}\\
    &\text{where} \enspace \frac{\partial \pi(\mathbf{u})}{\partial \mathbf{u}}=\begin{bmatrix}
    \nicefrac{f_x}{u_z} & 0 & -\nicefrac{f_x u_x}{u_z^2}\\
    0 & \nicefrac{f_y}{u_z} & -\nicefrac{f_y u_y}{u_z^2}\\
    \end{bmatrix}.
\end{align}
$\mathbf{I}$ is a $3\times3$ identity matrix, $(\cdot)^\wedge$ maps a vector $\in \mathbb{R}^3$ to its skew-symmetric matrix, $f_*$ are focal lengths and $u_*$ are coordinates of a 3D point $\mathbf{u}$. 

Using the current state of $\{\mathbf{p}, \mathbf{Y}\}$, each Gauss-Newton iteration establishes a linear system, that is solved for the update $\mathrm{\Delta}\mathbf{z}$ using LU decomposition: 
\begin{equation}
    \mathbf{J}^\top\mathbf{J} \mathrm{\Delta} \mathbf{z}=-\mathbf{J}^\top\mathbf{r}. 
    \label{eq:gn_update}
\end{equation}
We update the state in $T$ Gauss-Newton iterations and apply Jacobi preconditioning and a damping factor $\beta$ for stability. 
\subsection{End-to-End Training}
\label{ssec:end2end_training}
The whole pipeline, from the matching network to the pose optimization, is differentiable, which allows for a pose loss that guides the matching network to produce valuable matches and accurate confidences for robust pose optimization. The training objective $\mathcal{L}$ consists of a matching term $\mathcal{L}_{\mathrm{match}}$ \cite{Sarlin2020SuperGlueLF} and a pose term $\mathcal{L}_{\mathrm{pose}}$, which are balanced by the factor $\lambda$:
\begin{align}
    &\mathcal{L}=\sum_{(a,b)\in \mathcal{P}}\mathcal{L}_{\mathrm{match}}(a,b)+\lambda \mathcal{L}_{\mathrm{pose}}(a,b), \enspace \text{where} 
    \label{eq:total_loss} \\
    &\mathcal{L}_{\mathrm{match}}(a,b)=-\sum_{(i,j)\in \mathcal{T}_{ab}}\log \mathbf{P}_{ab,i,j} \\
    &\qquad-\sum_{i\in \mathcal{U}_{ab}}\log \mathbf{P}_{ab,i,\!\!\!\!\!\underbrace{\scriptstyle j_{\mathrm{max}}}_{\text{\scriptsize unmatched}}}-\sum_{j\in \mathcal{V}_{ab}}\log \mathbf{P}_{ab,\!\!\!\!\!\underbrace{\scriptstyle i_{\mathrm{max}}}_{\text{\scriptsize unmatched}}\!\!\!\!\!,j}, \nonumber \\
    &\mathcal{L}_{\mathrm{pose}}(a,b)=\cos^{-1}\left(\tfrac{\hat{\mathbf{t}}_{a\rightarrow b}\cdot\mathbf{t}_{a\rightarrow b}}{\left\Vert\hat{\mathbf{t}}_{a\rightarrow b}\right\Vert_2\cdot\left\Vert\mathbf{t}_{a\rightarrow b}\right\Vert_2}\right)\\
    &\qquad+\lambda_{\mathrm{rot}}\cos^{-1}\left(\tfrac{\mathrm{tr}\left(\hat{\mathbf{R}}_{a\rightarrow b}^\top\mathbf{R}_{a\rightarrow b}\right)-1}{2}\right). \nonumber
\end{align}
$\mathcal{L}_{\mathrm{match}}$ computes the negative log-likelihood of the assignment between an image pair. The labels are computed using the ground truth depth maps and camera parameters: $\mathcal{T}_{ab}$ is the set of matching keypoints, $\mathcal{U}_{ab}$ and $\mathcal{V}_{ab}$ identify unmatched keypoints from $I_a$ and $I_b$, respectively. 
$\mathcal{L}_{\mathrm{pose}}$ computes a transformation error between a pair of camera poses, where the translational and rotational components are balanced by $\lambda_{\mathrm{rot}}$. 
We found that training on the weighted eight-point result works equally well as training on both weighted eight-point and bundle adjustment, hence, $\mathcal{L}_{\mathrm{pose}}$ is applied on the weighted eight-point result.  
At test time, however, the pose refinement with bundle adjustment is highly beneficial as shown in the experiments (\cref{sec:results}). 
$\hat{\mathbf{R}}_{a\rightarrow b}$ and  $\hat{\mathbf{t}}_{a\rightarrow b}$ are the rotation matrix and translation vector of the estimated pose. $\mathbf{R}_{a\rightarrow b}$ and  $\mathbf{t}_{a\rightarrow b}$ define the ground truth transformation. 
We use the Adam optimizer \cite{Kingma2015AdamAM}. Further detail on the network architecture and training setup are provided in the supplementary material. 

\section{Results}
\label{sec:results}
We evaluate performance on indoor and outdoor pose estimation in a two-view and multi-view setting  (\cref{ssec:two_view_pose,ssec:multi_view_pose}) and runtime (\cref{ssec:runtime}). \cref{ssec:ablation} shows the effectiveness of end-to-end training and multi-view matching in an ablation study. A cross-dataset and matching evaluation is provided in the supplement. 

\paragraphNoSpace{Baselines.} Prior work, in particular SuperGlue~\cite{Sarlin2020SuperGlueLF}, has extensively demonstrated the superiority of the GNN approach over conventional matching. Hence, we focus on comparisons to recent matching networks: SuperGlue~\cite{Sarlin2020SuperGlueLF}, LoFTR~\cite{Sun2021LoFTRDL}, COTR~\cite{Jiang2021COTRCT}, and 3DG-STFM~\cite{Mao20223DGSTFM}. We additionally compare to a non-learning-based matcher, i.e., mutual nearest neighbor search on the SuperPoint \cite{DeTone2018SuperPointSI} descriptors. This serves to confirm the effectiveness of SuperGlue and our method, which both use SuperPoint descriptors.
\subsection{Two-View Pose Estimation}
\label{ssec:two_view_pose}
\begin{table}[tb]
\begin{center}
\resizebox{\linewidth}{!}{
  \begin{tabular}{l >{\centering\arraybackslash}p{1.25cm} >{\centering\arraybackslash}p{0.8cm} >{\centering\arraybackslash}p{0.8cm} >{\centering\arraybackslash}p{0.8cm}}
    \toprule
    &\multirow{2}{\linewidth}{\centering Pose est.\ method}& \multicolumn{3}{c}{Pose error AUC [\%] $\uparrow$} \\
    \cmidrule(lr){3-5}
    && @5\degree & @10\degree & @20\degree \\
    \midrule
    Mutual nearest neighbor & \multirow{6}{*}{\rotatebox[origin=c]{90}{RANSAC}} & 9.5 & 21.6 & 35.7 \\
    SuperGlue \cite{Sarlin2020SuperGlueLF} & & 16.2 & 33.8 & 51.8 \\
    LoFTR \cite{Sun2021LoFTRDL} & & 22.1 & 40.8 & 57.6 \\
    COTR \cite{Jiang2021COTRCT} cross-dataset & & 11.8 & 26.5 & 42.5 \\
    3DG-STFM \cite{Mao20223DGSTFM} & & 23.6 & 43.6 & 61.2 \\
    Ours w/o multi-view & & 20.7 & 41.3 & 60.7 \\
    \cmidrule(lr){2-2}
    Mutual nearest neighbor & \multirow{6}{*}{\rotatebox[origin=c]{90}{Weight.\ 8-point}} & 0.0 & 0.1 & 0.7 \\
    SuperGlue \cite{Sarlin2020SuperGlueLF} & & 11.7 & 26.8 & 45.6 \\
    LoFTR \cite{Sun2021LoFTRDL} & & 15.0 & 30.6 & 47.3 \\
    COTR \cite{Jiang2021COTRCT} cross-dataset & & 3.2 & 9.5 & 20.2 \\
    3DG-STFM \cite{Mao20223DGSTFM} & & 10.1 & 23.4 & 39.5 \\
    Ours w/o multi-view & & 20.7 & 41.6 & 61.7 \\
    \cmidrule(lr){2-2}
    Mutual nearest neighbor & \multirow{6}{*}{\rotatebox[origin=c]{90}{\parbox{2.5cm}{\centering RANSAC + bundle adjust.}}} & 10.1 & 22.4 & 36.3 \\
    SuperGlue \cite{Sarlin2020SuperGlueLF} & & 17.0 & 35.2 & 54.0 \\
    LoFTR \cite{Sun2021LoFTRDL} & & 22.4 & 41.0 & 57.7 \\
    COTR \cite{Jiang2021COTRCT} cross-dataset & & 12.6 & 27.7 & 43.5 \\
    3DG-STFM \cite{Mao20223DGSTFM} & & 23.3 & 42.4 & 59.1 \\
    Ours w/o multi-view & & 23.1 & 43.6 & 62.3 \\
    \cmidrule(lr){2-2}
    Mutual nearest neighbor & \multirow{6}{*}{\rotatebox[origin=c]{90}{\parbox{2.5cm}{\centering Weight.\ 8-point + bundle adjust.}}}  & 0.0 & 0.3 & 1.8 \\
    SuperGlue \cite{Sarlin2020SuperGlueLF} & & 20.6 & 40.0 & 58.7 \\
    LoFTR \cite{Sun2021LoFTRDL} & & 24.0 & 42.8 & 59.1 \\
    COTR \cite{Jiang2021COTRCT} cross-dataset & & 8.5 & 19.6 & 33.9 \\
    3DG-STFM \cite{Mao20223DGSTFM} & & 20.3 & 37.9 & 54.1 \\
    Ours w/o multi-view & & \textbf{25.7} & \textbf{47.2} & \textbf{66.4} \\
    \bottomrule
  \end{tabular}
  }
\end{center}
\caption{Baseline comparison on two-view, wide-baseline, indoor pose estimation on ScanNet. Through end-to-end training with pose optimization, our network learns to predict valuable matches for pose estimation, and downweights outliers. This enables accurate weighted pose estimation, which outperforms the baselines. ``cross-dataset'' indicates that COTR was trained on MegaDepth.}
\label{tab:two_view_pose_scannet}
\end{table}
\begin{table}[tb]
\begin{center}
\resizebox{\linewidth}{!}{
  \begin{tabular}{l >{\centering\arraybackslash}p{1.25cm} >{\centering\arraybackslash}p{0.8cm} >{\centering\arraybackslash}p{0.8cm} >{\centering\arraybackslash}p{0.8cm}}
    \toprule
    &\multirow{2}{\linewidth}{\centering Pose est.\ method}& \multicolumn{3}{c}{Pose error AUC [\%] $\uparrow$} \\
    \cmidrule(lr){3-5}
    && @5\degree & @10\degree & @20\degree \\
    \midrule
    Mutual nearest neighbor & \multirow{6}{*}{\rotatebox[origin=c]{90}{RANSAC}} & 32.2 & 47.6 & 55.2 \\
    SuperGlue \cite{Sarlin2020SuperGlueLF} & & 43.4 & 61.6 & 76.2 \\
    LoFTR \cite{Sun2021LoFTRDL} & & 52.8 & 69.2 & 81.2 \\
    COTR \cite{Jiang2021COTRCT} & & 35.2 & 53.9 & 69.6 \\
    3DG-STFM \cite{Mao20223DGSTFM} & & 52.6 & 68.5 & 80.0 \\
    Ours w/o multi-view & & 49.5 & 66.7 & 79.9 \\
    \cmidrule(lr){2-2}
    Mutual nearest neighbor & \multirow{6}{*}{\rotatebox[origin=c]{90}{Weight.\ 8-point}} & 0.1 & 0.2 & 1.0 \\
    SuperGlue \cite{Sarlin2020SuperGlueLF} & & 23.8 & 36.2 & 49.2 \\
    LoFTR \cite{Sun2021LoFTRDL} & & 15.5 & 27.1 & 41.6 \\
    COTR \cite{Jiang2021COTRCT} & & 29.6 & 43.4 & 57.2 \\
    3DG-STFM \cite{Mao20223DGSTFM} & & 4.0 & 9.5 & 19.8 \\
    Ours w/o multi-view & & 46.9 & 62.8 & 76.3 \\
    \cmidrule(lr){2-2}
    Mutual nearest neighbor & \multirow{6}{*}{\rotatebox[origin=c]{90}{\parbox{2.5cm}{\centering RANSAC + bundle adjust.}}} & 34.9 & 49.5 & 61.9 \\
    SuperGlue \cite{Sarlin2020SuperGlueLF} & & 48.3 & 65.2 & 78.3 \\
    LoFTR \cite{Sun2021LoFTRDL} & & 52.8 & 69.6 & 82.0 \\
    COTR \cite{Jiang2021COTRCT} cross-dataset & & 45.0 & 61.1 & 73.8 \\
    3DG-STFM \cite{Mao20223DGSTFM} & & 51.2 & 67.7 & 80.2 \\
    Ours w/o multi-view & & 55.3 & 70.8 & 82.3 \\
    \cmidrule(lr){2-2}
    Mutual nearest neighbor & \multirow{6}{*}{\rotatebox[origin=c]{90}{\parbox{2.5cm}{\centering Weight.\ 8-point + bundle adjust.}}}  & 0.1 & 0.8 & 4.3 \\
    SuperGlue \cite{Sarlin2020SuperGlueLF} & & 40.3 & 53.6 & 65.6 \\
    LoFTR \cite{Sun2021LoFTRDL} & & 25.7 & 40.0 & 54.7 \\
    COTR \cite{Jiang2021COTRCT} & & 47.1 & 61.3 & 72.5 \\
    3DG-STFM \cite{Mao20223DGSTFM} & & 10.2 & 20.0 & 35.0 \\
    Ours w/o multi-view & & \textbf{61.2} & \textbf{74.9} & \textbf{85.0} \\
    \bottomrule
  \end{tabular}
  }
\end{center}
\caption{Baseline comparison on two-view, wide-baseline, outdoor pose estimation on MegaDepth. The pose optimization objective guides our method to produce matches with accurate confidences for weighted pose estimation, leading to higher pose accuracy than the baselines relying on RANSAC.}
\label{tab:two_view_pose_megadepth}
\end{table}
Following prior work \cite{Sarlin2020SuperGlueLF,Sun2021LoFTRDL,Mao20223DGSTFM}, we evaluate on the same 1500 image pairs of ScanNet and MegaDepth and compute the area under the curve (AUC) in \% at the thresholds $[5\degree, 10\degree, 20\degree]$ of the pose error, i.e., the maximum of rotation and translation error, where the translation error is 
the angle between translation vectors, since poses are only determined up to an unknown scale factor.
\cref{tab:two_view_pose_scannet,tab:two_view_pose_megadepth} list the AUC metrics for four pose estimation methods: \begin{enumerate*}[label=(\roman*)] \item essential matrix estimation with RANSAC,  \item weighted eight-point algorithm (\cref{ssec:pose_optimization}), \item RANSAC followed by $T=10$ bundle adjustment iterations (\cref{ssec:pose_optimization}) and \item weighted eight-point algorithm followed by $T=10$ bundle adjustment iterations (\cref{ssec:pose_optimization}) \end{enumerate*}. 
The results show that our method outperforms the baselines on two-view pose estimation. For our method, the combination of weighted eight-point algorithm and bundle adjustment is stronger than pose estimation with RANSAC in the indoor and outdoor setting. This shows that end-to-end training enables the learning of accurate confidences that down-weight outliers and render RANSAC unnecessary. 
\subsection{Multi-View Pose Estimation}
\label{ssec:multi_view_pose}
\begin{table}[tb]
\begin{center}
\resizebox{\linewidth}{!}{
  \begin{tabular}{l >{\centering\arraybackslash}p{0.8cm} >{\centering\arraybackslash}p{0.8cm} >{\centering\arraybackslash}p{0.8cm} >{\centering\arraybackslash}p{0.8cm} >{\centering\arraybackslash}p{0.8cm} >{\centering\arraybackslash}p{0.8cm}}
    \toprule
    & \multicolumn{3}{c}{Transl.\ error AUC [\%] $\uparrow$} & \multicolumn{3}{c}{Rot.\ error AUC [\%] $\uparrow$} \\
    \cmidrule(lr){2-4}
    \cmidrule(lr){5-7}
    & @5\degree & @10\degree & @20\degree & @5\degree & @10\degree & @20\degree\\
    \midrule
    Mutual nearest neighbor & 8.5 & 17.8 & 31.0 & 33.0 & 48.4 & 62.8 \\
    SuperGlue \cite{Sarlin2020SuperGlueLF} & 21.3 & 37.5 & 53.7 & 54.2 & 71.0 & 82.6 \\
    LoFTR \cite{Sun2021LoFTRDL} & 20.6 & 36.9 & 53.7 & 57.3 & 72.0 & 82.0 \\
    COTR \cite{Jiang2021COTRCT} cross-dataset & 10.9 & 22.4 & 36.9 & 38.8 & 53.6 & 66.3 \\
    3DG-STFM \cite{Mao20223DGSTFM} & 22.0 & 38.7 & 55.5 & 57.0 & 72.7 & 83.0 \\
    Ours & \textbf{26.9} & \textbf{45.6} & \textbf{63.0} & \textbf{64.2} & \textbf{78.8} & \textbf{87.7} \\
    \bottomrule
  \end{tabular}
  }
\end{center}
\caption{Baseline comparison on multi-view indoor pose estimation on ScanNet. Our multi-view and end-to-end approach, predicts matches and confidences that improve pose estimation compared to the pairwise baselines. ``cross-dataset'' indicates that COTR was trained on MegaDepth.}
\label{tab:multi_view_pose_scannet}
\begin{center}
\resizebox{\linewidth}{!}{
  \begin{tabular}{l >{\centering\arraybackslash}p{0.8cm} >{\centering\arraybackslash}p{0.8cm} >{\centering\arraybackslash}p{0.8cm} >{\centering\arraybackslash}p{0.8cm} >{\centering\arraybackslash}p{0.8cm} >{\centering\arraybackslash}p{0.8cm}}
    \toprule
    & \multicolumn{3}{c}{Transl.\ error AUC [\%] $\uparrow$} & \multicolumn{3}{c}{Rot.\ error AUC [\%] $\uparrow$} \\
    \cmidrule(lr){2-4}
    \cmidrule(lr){5-7}
    & @5\degree & @10\degree & @20\degree & @5\degree & @10\degree & @20\degree\\
    \midrule
    Mutual nearest neighbor & 2.8 & 5.6 & 10.6 & 3.3 & 6.6 & 12.3 \\
    SuperGlue \cite{Sarlin2020SuperGlueLF} & 17.1 & 24.0 & 32.7 & 17.9 & 25.9 & 35.3 \\
    Ours w/o multi-view & 19.4 & 27.8 & 38.4 & 20.9 & 30.5 & 41.8 \\
    Ours w/o end-to-end & 28.5 & 35.4 & 42.7 & 29.4 & 38.0 & 46.2 \\
    Ours & \textbf{33.2} & \textbf{42.1} & \textbf{51.6} & \textbf{35.1} & \textbf{45.8} & \textbf{56.2} \\
    \bottomrule
  \end{tabular}
  }
\end{center}
\caption{Baseline comparison and ablation study on multi-view indoor pose estimation on Matterport3D. The full version of our method, with multi-view matching and end-to-end training with pose optimization, achieves best performance.}
\label{tab:multi_view_pose_matterport}
\begin{center}
\resizebox{\linewidth}{!}{
  \begin{tabular}{l >{\centering\arraybackslash}p{0.8cm} >{\centering\arraybackslash}p{0.8cm} >{\centering\arraybackslash}p{0.8cm} >{\centering\arraybackslash}p{0.8cm} >{\centering\arraybackslash}p{0.8cm} >{\centering\arraybackslash}p{0.8cm}}
    \toprule
    & \multicolumn{3}{c}{Transl.\ error AUC [\%] $\uparrow$} & \multicolumn{3}{c}{Rot.\ error AUC [\%] $\uparrow$} \\
    \cmidrule(lr){2-4}
    \cmidrule(lr){5-7}
    & @5\degree & @10\degree & @20\degree & @5\degree & @10\degree & @20\degree\\
    \midrule
    Mutual nearest neighbor & 12.0 & 20.1 & 31.9 & 23.4 & 36.7 & 51.8 \\
    SuperGlue \cite{Sarlin2020SuperGlueLF} & 47.3 & 58.7 & 68.9 & 60.9 & 73.6 & 83.4 \\
    LoFTR \cite{Sun2021LoFTRDL} & 48.7 & 59.5 & 69.5 & 63.9 & 75.3 & 84.0 \\
    COTR \cite{Jiang2021COTRCT} & 37.9 & 48.1 & 58.3 & 49.8 & 61.9 & 72.7 \\
    3DG-STFM \cite{Mao20223DGSTFM} & 44.5 & 55.3 & 65.8 & 59.5 & 71.9 & 81.7 \\
    Ours & \textbf{52.1} & \textbf{63.0} & \textbf{72.5} & \textbf{66.7} & \textbf{77.8} & \textbf{85.9} \\
    \bottomrule
  \end{tabular}
  }
\end{center}
\caption{Baseline comparison on multi-view outdoor pose estimation on MegaDepth. Through multi-view matching and end-to-end training, our method achieves higher pose estimation accuracy than the baselines.}
\label{tab:multi_view_pose_megadepth}
\end{table}
For multi-view evaluation, we sample test images with the same overlap criterion as used by prior work to sample image pairs \cite{Sarlin2020SuperGlueLF,Sun2021LoFTRDL,Mao20223DGSTFM}. However, instead of sampling a pair, we sample a 5-tuple, by appending three more images that each satisfy the overlap criterion to the previous one. Further detail and overlap ranges are provided in the supplement. 
Besides ScanNet and MegaDepth, we evaluate on Matterport3D, which is particularly challenging for matching, as view captures are much more sparse, i.e., neighboring images are $60\degree$ horizontally and $30\degree$ vertically apart. This difficult dataset, serves to measure robustness on the pose estimation task. 

Multi-view pose estimation is evaluated as follows: \begin{enumerate*}[label=(\roman*)] \item Feature matches are computed. Baselines that operate on image pairs are run on all possible pairs of the tuple. \item Relative poses are estimated between all possible pairs using the best performing two-view pose estimation from \cref{ssec:two_view_pose}. \item Absolute poses are determined through robust estimators for rotation \cite{ChatterjeeRotationAveraging} and translation \cite{zyesil2015RobustCL}, which take initial absolute poses and relative poses as input. The initial absolute poses are obtained by composing relative poses along edges of a maximum spanning tree on the match graph, where edge weights are inlier counts from the previous step. 
\item Bundle adjustment jointly optimizes all poses by minimizing the confidence-weighted reprojection error of inlier matches using Ceres Solver for non-linear least squares optimization \cite{Agarwal_Ceres_Solver_2022}\end{enumerate*}. 
The pose estimation performance is measured by the translation and rotation error AUC between all possible pairs of the tuple. 

The quantitative results (\cref{tab:multi_view_pose_scannet,tab:multi_view_pose_matterport,tab:multi_view_pose_megadepth}) show that our method achieves higher AUC metrics than the baselines across all thresholds in the indoor and outdoor setting. The metrics on Matterport3D are overall lower than on ScanNet and MegaDepth, due to the smaller overlap between images. 
In this scenario, our method outperforms SuperGlue with a larger gap than on ScanNet or MegaDepth, which shows that our approach copes better with the more challenging setting in Matterport3D. 
For qualitative comparison, we visualize the reprojection error by projecting the ground truth depth maps from all other views using the estimated poses, scaled according to the ground truth (\cref{fig:scannet_results,fig:megadepth_results}). 
With multi-view reasoning during matching and learned outlier rejection through end-to-end training, our method is robust to challenging situations, like repetitive patterns (\cref{fig:scannet_results} sample 2) 
or large viewpoint changes (\cref{fig:scannet_results} sample 1). 
\begin{figure*}[tb]
\begin{center}
    \begin{tikzpicture}[squarednode/.style={rectangle, draw=none, fill=white, very thin, minimum size=2mm, text opacity=1,fill opacity=0.}]
        \node[anchor=south west,inner sep=0] (image) at (0,0)
        {\includegraphics[width=\linewidth]{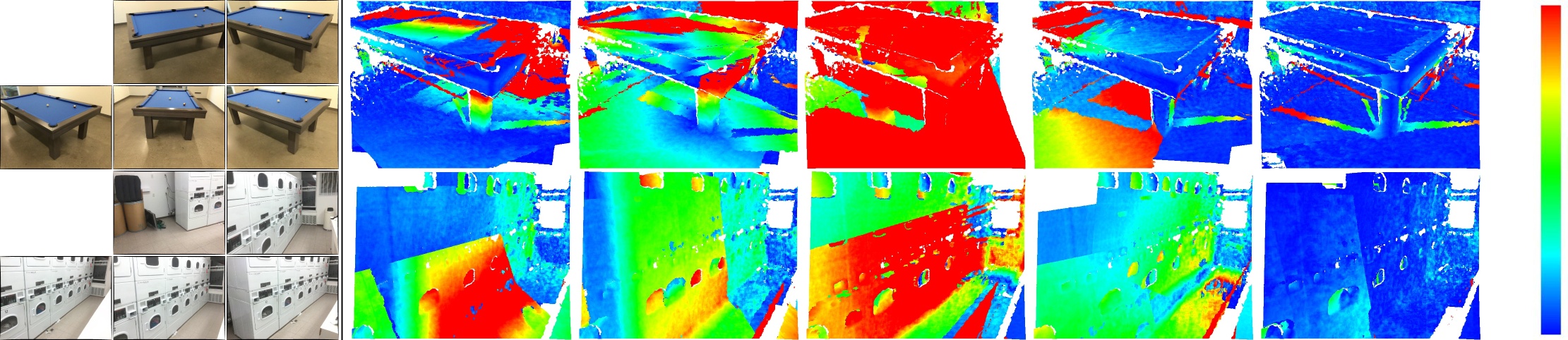}};
        \begin{scope}[x=(image.south east),y=(image.north west)]
        \node[squarednode] at (0.035, 0.89) (a) {\bf 1};
        \node[squarednode] at (0.035, 0.38) (b) {\bf 2};
        \node[squarednode] at (0.11, -0.06) (g) {Input 5-tuples};
        \node[squarednode] at (0.295, -0.06) (g) {SuperGlue \cite{Sarlin2020SuperGlueLF}};
        \node[squarednode] at (0.44, -0.06) (g) {LoFTR \cite{Sun2021LoFTRDL}};
        \node[squarednode] at (0.585, -0.06) (g) {COTR \cite{Jiang2021COTRCT}};
        \node[squarednode] at (0.73, -0.06) (g) {3DG-STFM \cite{Mao20223DGSTFM}};
        \node[squarednode] at (0.87, -0.06) (g) {Ours};
        \node[squarednode] at (0.965, 0.983) (g) {1m};
        \node[squarednode] at (0.965, 0.02) (g) {0m};
        \end{scope}
    \end{tikzpicture}
\end{center}
\caption{Reprojection error (right) for estimated camera poses on ScanNet 5-tuples (left). With multi-view matching and end-to-end training, our method successfully handles challenging pose estimation scenarios, while baselines have severe camera pose errors.}
\label{fig:scannet_results}
\end{figure*}
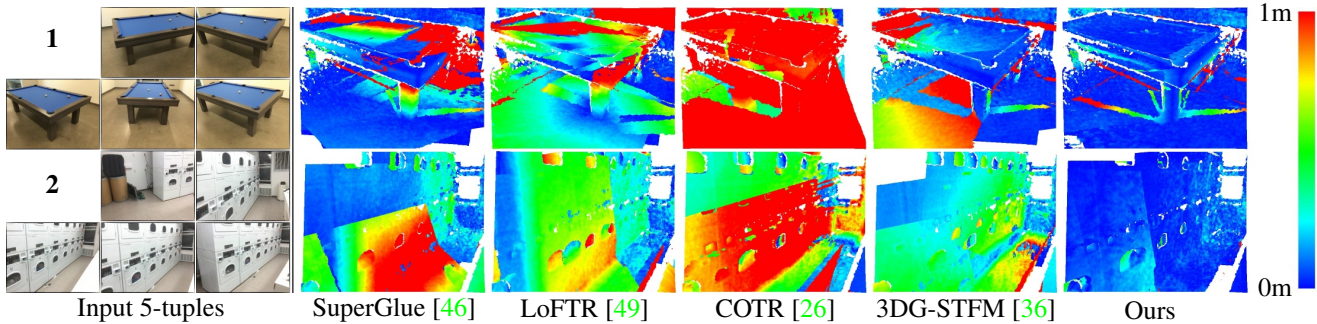
\begin{figure*}[tb]
\begin{center}
    \begin{tikzpicture}[squarednode/.style={rectangle, draw=none, fill=white, very thin, minimum size=2mm, text opacity=1,fill opacity=0.}]
        \node[anchor=south west,inner sep=0] (image) at (0,0)
        {\includegraphics[width=\linewidth]{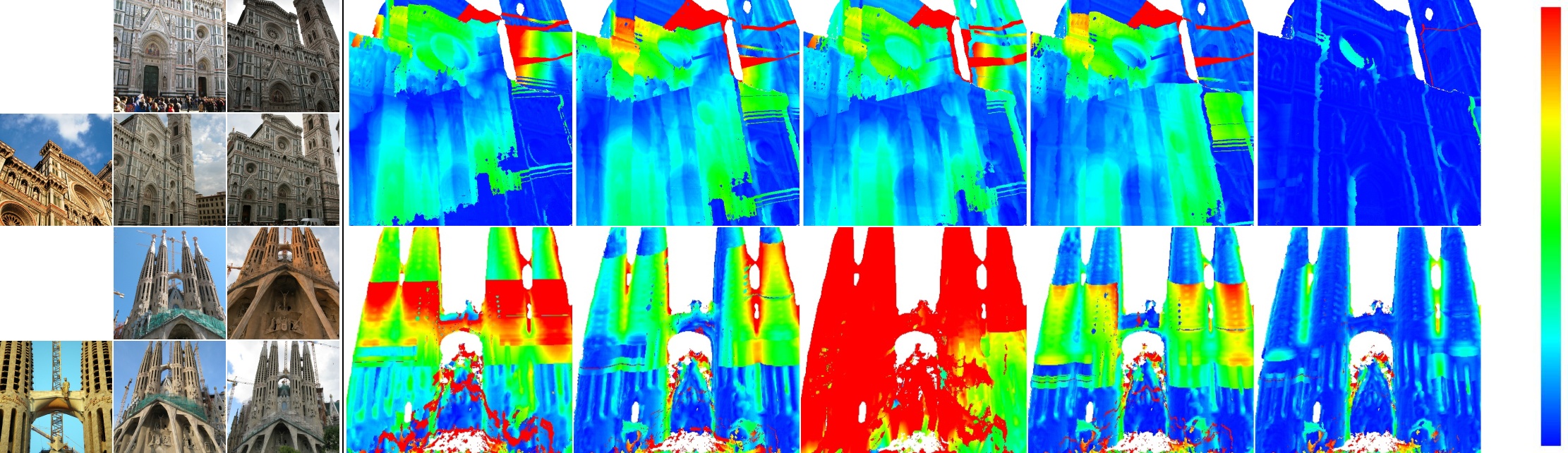}};
        \begin{scope}[x=(image.south east),y=(image.north west)]
        \node[squarednode] at (0.035, 0.89) (a) {\bf 1};
        \node[squarednode] at (0.035, 0.38) (b) {\bf 2};
        \node[squarednode] at (0.11, -0.05) (g) {Input 5-tuples};
        \node[squarednode] at (0.29, -0.05) (g) {SuperGlue \cite{Sarlin2020SuperGlueLF}};
        \node[squarednode] at (0.44, -0.05) (g) {LoFTR \cite{Sun2021LoFTRDL}};
        \node[squarednode] at (0.58, -0.05) (g) {COTR \cite{Jiang2021COTRCT}};
        \node[squarednode] at (0.725, -0.05) (g) {3DG-STFM \cite{Mao20223DGSTFM}};
        \node[squarednode] at (0.87, -0.05) (g) {Ours};
        \node[squarednode] at (0.965, 0.983) (g) {1};
        \node[squarednode] at (0.965, 0.02) (g) {0};
        \end{scope}
    \end{tikzpicture}
\end{center}
\caption{Reprojection error (right) for estimated camera poses on MegaDepth 5-tuples (left). Through multi-view matching and end-to-end training, our method successfully estimates camera poses in challenging outdoor scenarios, while baselines show misalignment. Reprojection errors are visualized in the MegaDepth scaling.}
\label{fig:megadepth_results}
\end{figure*}
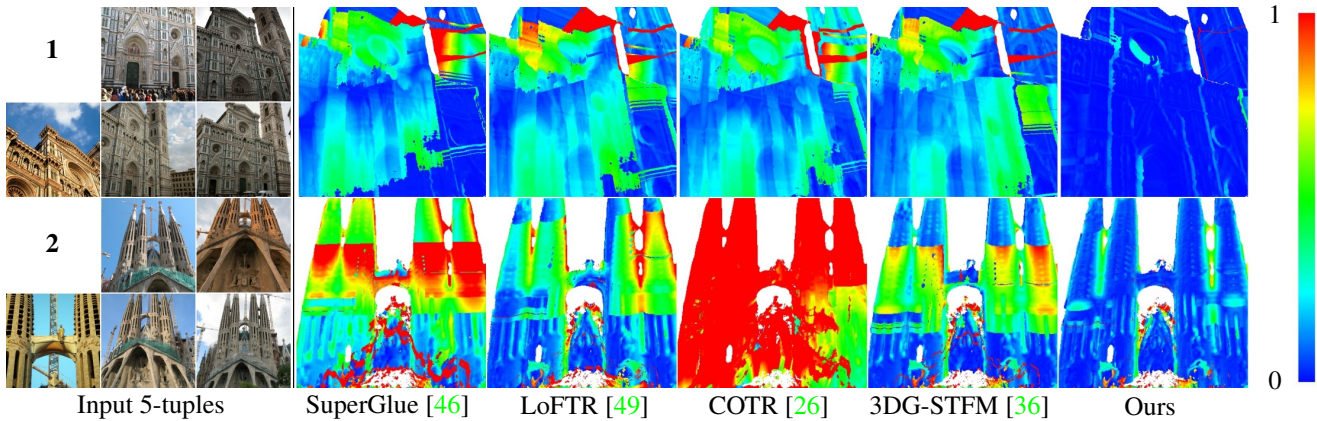

We further evaluate multi-view pose estimation using the protocol of the Image Matching Challenge (IMC) 2021~\cite{jin2021image}. It provides a multi-view setting, where COLMAP~\cite{schoenberger2016sfm} Structure-from-Motion (SfM) estimates camera poses on groups of 5-25 internet images of tourist attractions. 
\cref{tab:imc_multi_view} lists the pose error AUC metrics for the detector-based methods, SuperGlue and Ours. Even though COLMAP does not consider our learned confidence weights, we observe a clear improvement through our end-to-end and multi-view approach. 
\begin{table}[tb]
\begin{center}
\resizebox{0.65\linewidth}{!}{
\begin{tabular}{lcc}
\toprule
& \multicolumn{2}{c}{Pose error AUC [\%] $\uparrow$} \\
\cmidrule(lr){2-3}
& @5\degree & @10\degree \\
\midrule
SuperGlue \cite{Sarlin2020SuperGlueLF} & 70.0 & 80.2 \\
Ours & \textbf{74.5} & \textbf{83.4} \\
\bottomrule
\end{tabular}
}
\end{center}
\caption{IMC multi-view evaluation using COLMAP SfM on the PhotoTourism dataset. Although COLMAP does not use matching confidences, there is a clear benefit from our multi-view matching method.}
\label{tab:imc_multi_view}
\end{table}

Details on the baseline comparisons, further qualitative results and a cross-attention visualization are provided in the supplementary material. 

\begin{figure*}[tb]
\begin{center}
    \begin{tikzpicture}[squarednode/.style={rectangle, draw=none, fill=white, very thin, minimum size=2mm, text opacity=1,fill opacity=0.}]
        \node[anchor=south west,inner sep=0] (image) at (0,0)
        {\includegraphics[width=\linewidth]{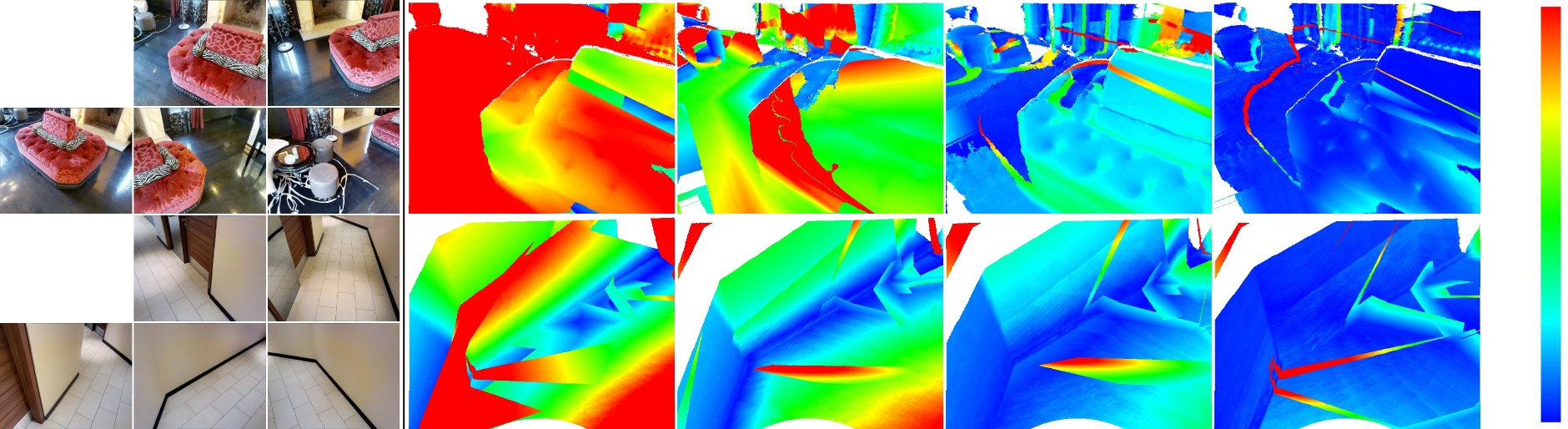}};
        \begin{scope}[x=(image.south east),y=(image.north west)]
        \node[squarednode] at (0.042, 0.885) (a) {\bf 1};
        \node[squarednode] at (0.042, 0.375) (b) {\bf 2};
        \node[squarednode] at (0.128, -0.05) (g) {Input 5-tuples};
        \node[squarednode] at (0.345, -0.05) (g) {SuperGlue \cite{Sarlin2020SuperGlueLF}};
        \node[squarednode] at (0.515, -0.05) (g) {Ours w/o multi-view};
        \node[squarednode] at (0.69, -0.05) (g) {Ours w/o end-to-end};
        \node[squarednode] at (0.86, -0.05) (g) {Ours};
        \node[squarednode] at (0.965, 0.983) (g) {1m};
        \node[squarednode] at (0.965, 0.02) (g) {0m};
        \end{scope}
    \end{tikzpicture}
\end{center}
\caption{Reprojection error (right) for estimated camera poses on Matterport3D 5-tuples (left). Our complete method improves camera alignment over the ablated versions and SuperGlue, showing the importance of multi-view matching and end-to-end training.}
\label{fig:matterport_results}
\end{figure*}
\subsection{Runtime}
\label{ssec:runtime}
\cref{tab:runtime} compares runtime for matching and pose estimation. Our method requires the same amount of time as SuperGlue for matching an image pair, however, we reduce runtime by 9\% when matching a 5-tuple. The savings stem from fewer intra-frame GNN messages in multi-view matching compared to  matching the corresponding pairs individually (see supplementary material). The detector-free baselines take far more time for matching. 
Our method more than halves the RANSAC time compared to SuperGlue. This shows that our confidences allow for better outlier pre-filtering by confidence thresholding, which improves the ratio between inliers and outliers prior to RANSAC. 
Our proposed weighted pose estimation (weighted eight-point + bundle adjustment)---besides reducing the pose error (\cref{ssec:two_view_pose})---reduces the runtime on SuperGlue matches and our matches by half, compared to RANSAC on SuperGlue matches. Only COTR, due to a smaller number of matches, has a shorter pose estimation runtime, however, its matching time is multiple orders of magnitude higher and the pose  accuracy is lower. 
All runtime is measured on a Nvidia GeForce RTX 2080. For a fair comparison to the detector-free matchers, the matching time of SuperGlue and our method includes the SuperPoint inference time. 
\begin{table}[htb]
\begin{center}
\resizebox{\linewidth}{!}{
  \begin{tabular}{l >{\raggedleft\arraybackslash}p{1.4cm} >{\raggedleft\arraybackslash}p{1.6cm} >{\raggedleft\arraybackslash}p{1.2cm}  >{\raggedleft\arraybackslash}p{1.1cm} >{\raggedleft\arraybackslash}p{1.1cm}}
    \toprule
    & \multicolumn{2}{c}{Matching time $\downarrow$} & \multicolumn{3}{c}{Pose estimation time $\downarrow$} \\
    \cmidrule(lr){2-3}
    \cmidrule(lr){4-6}
    & 2-view & 5-view & RANSAC & Weight. & Bundle \\
    &  & $\widehat{=}$ 10 pairs & & 8-point & adjust. \\
    \midrule
    SuperGlue \cite{Sarlin2020SuperGlueLF}& \textbf{60}\,ms & 371\,ms & 126\,ms & \textbf{5}\,ms & 56\,ms \\
    LoFTR \cite{Sun2021LoFTRDL} & 108\,ms & 976\,ms & 148\,ms & 9\,ms & 511\,ms \\
    COTR \cite{Jiang2021COTRCT} & 37950\,ms & 357096\,ms & 126\,ms & \textbf{5}\,ms & \textbf{47}\,ms \\
    3DG-STFM \cite{Mao20223DGSTFM} & 130\,ms & 1176\,ms & 201\,ms & 10\,ms & 735\,ms \\
    Ours & \textbf{60}\,ms & \textbf{338}\,ms & \textbf{52}\,ms & \textbf{5}\,ms & 56\,ms \\
    \bottomrule
  \end{tabular}
  }
\end{center}
\caption{Matching and pose estimation time on ScanNet. Multi-view matching is faster than matching the corresponding pairs. Our confidences enable effective thresholding prior to RANSAC, reducing its runtime. 
Weighted eight-point + bundle adjustment is faster or comparable to RANSAC on SuperGlue and our matches.}
\label{tab:runtime}
\end{table}
\subsection{Ablation Study}
\label{ssec:ablation}
The quantitative results on Matterport3D (\cref{tab:multi_view_pose_matterport}) show that the full version of our method achieves the best performance. This is consistent with the qualitative results (\cref{fig:matterport_results}), as well as the ablation results on ScanNet and MegaDepth, which are provided in the supplement. 

\paragraphNoSpace{Without Multi-View.}
Omitting multi-view in the GNN causes an average performance drop of 14.2\% on Matterport3D.
This suggests that the multi-view receptive field supports information flow from other views to bridge gaps, where the overlap is small. Sample 1 in \cref{fig:matterport_results} shows that without multi-view reasoning, the matching fails to resolve large viewpoint changes and difficult object symmetries. 

\paragraphNoSpace{Without End-to-End.}
Without end-to-end training the average performance drops by 7.3\%. 
This shows that end-to-end training enables the learning of an outlier down-weighting, that improves pose estimation. Dropping end-to-end leads to increased misalignment in \cref{fig:matterport_results}. 

\begin{figure}[htb]
\begin{center}
\begin{tikzpicture}[squarednode/.style={rectangle, draw=white, fill=white, very thin, minimum size=2mm, text opacity=1,fill opacity=0,draw opacity=0}]
        \node[anchor=south west,inner sep=0] (image) at (0,0)
        {\includegraphics[width=\linewidth,trim={-0.3cm 0cm 0cm 0cm},clip]{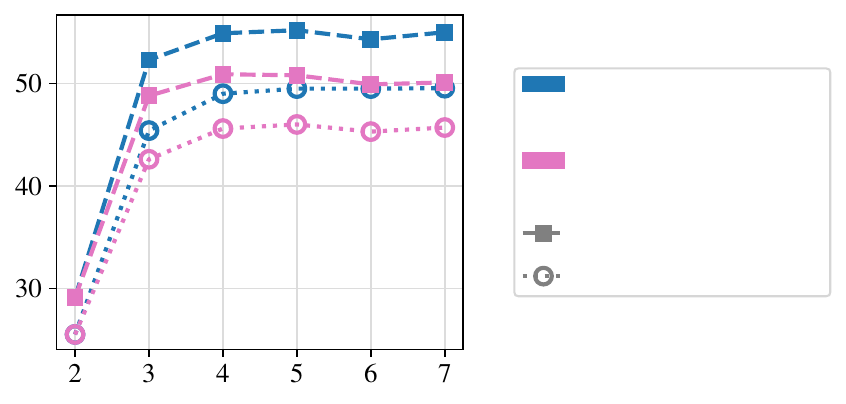}};
        \begin{scope}[x=(image.south east),y=(image.north west)]
        \small
        \node[squarednode,rotate=90] at (0., 0.55) (b) {Pose error\ AUC@20\degree [\%]};
        \node[squarednode] at (0.34, -0.03) (a) {Number of images};
        \node[squarednode] at (0.81, 0.78) (a) {Joint matching};
        \node[squarednode] at (0.815, 0.7) (a) {(w/ multi-view)};
        \node[squarednode] at (0.835, 0.6) (a) {Pairwise matching};
        \node[squarednode]  at (0.825, 0.52) (a) {(w/o multi-view)};
        \node[squarednode] at (0.805, 0.425) (a) {w/ end-to-end};
        \node[squarednode] at (0.815, 0.325) (a) {w/o end-to-end};
        \end{scope}
\end{tikzpicture}
\end{center}
   \caption{Pose error AUC on sequences of 9 images on ScanNet using variable number of images in pairwise or joint matching. Multi-view matching across $\sim$5 images combined with end-to-end training gives the best performance.}
   \label{fig:number_images}
\end{figure}
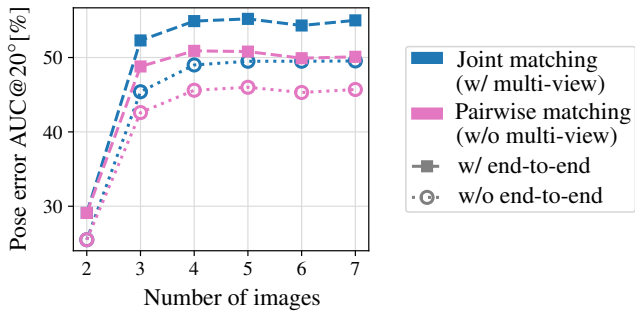

\paragraphNoSpace{Variable Number of Input Views.} In \cref{fig:number_images}, we investigate the impact of the number of images used for matching, both in pairwise (w/o multi-view) and joint (w/ multi-view) manner. 
The experiment is conducted on sequences of 9 images which are generated on ScanNet as described in \cref{ssec:multi_view_pose}. 
The results show that pose estimation improves, when matching across a larger span of neighboring images. The curves, however, plateau when a larger window size does not bring any more relevant images into the matching. 
Additionally, the results show the benefit of joint matching in a single graph as opposed to matching all possible image pairs individually. 

\paragraphNoSpace{Variable Image Overlap.} Evaluations on reduced image overlap are provided in the supplementary material. 
\subsection{Limitations}
One of our contributions is the end-to-end differentiablity of the pose optimization that guides the matching network.
While this significantly improves the pose estimation results, we currently only backpropgate gradients to the matching network, but do not update keypoint descriptors; i.e., we use existing SuperPoint~\cite{DeTone2018SuperPointSI}.
However, we believe that jointly training feature descriptors is a promising avenue to even further improve performance. Besides, more recent keypoint detectors and descriptors like ASLFeat~\cite{luo2020aslfeat}, in contrast to SuperPoint, provide subpixel accuracy, which can boost subsequent matching and pose estimation. 

\section{Conclusion}
We have presented a method that couples multi-view feature matching and pose optimization into an end-to-end trainable pipeline. 
Our graph neural network matches features across multiple views in a joint fashion, which enables globally informed matching solutions. 
Combined with differentiable pose optimization, gradients inform the matching network, which learns to reject outliers to produce valuable matches for pose estimation. 
Our method significantly improves pose estimation compared to prior work. 
In particular, we observe increased robustness in challenging settings, such as in presence of repetitive structure or small image overlap as in the Matterport3D dataset. 
Overall, we believe that our end-to-end approach is an important stepping stone towards an end-to-end trained SLAM method. 
\section*{Acknowledgements}
This work was supported by the ERC Starting Grant Scan2CAD (804724), the German Research Foundation (DFG) Grant ``Making Machine Learning on Static and Dynamic 3D Data Practical'', and the German Research Foundation (DFG) Research Unit ``Learning and Simulation in Visual Computing''. We thank Angela Dai for the video voice over. 

{\small
\bibliographystyle{ieee_fullname}
\bibliography{egbib}
}
\clearpage
\appendix
\section{Ablation Study}
\label{sec:ablation_suppl}
\paragraphNoSpace{Multi-View \& End-to-End.} The quantitative ablation results on ScanNet~\cite{Dai2017ScanNetR3} and MegaDepth~\cite{Li2018MegaDepthLS} confirm that the full version of our method achieves highest performance (\cref{tab:ablation_scannet,tab:ablation_megadepth}). 
\cref{fig:matterport_results_suppl} shows qualitative results of the ablation experiments on Matterport3D~\cite{Chang2017Matterport3DLF}. Clearly, multi-view matching and end-to-end training support the correspondence reasoning and improve camera alignment, despite the extreme viewpoint changes. 
\begin{table}[htb]
\begin{center}
\resizebox{\linewidth}{!}{
  \begin{tabular}{l >{\centering\arraybackslash}p{0.8cm} >{\centering\arraybackslash}p{0.8cm} >{\centering\arraybackslash}p{0.8cm} >{\centering\arraybackslash}p{0.8cm} >{\centering\arraybackslash}p{0.8cm} >{\centering\arraybackslash}p{0.8cm}}
    \toprule
    & \multicolumn{3}{c}{Transl.\ error AUC [\%] $\uparrow$} & \multicolumn{3}{c}{Rot.\ error AUC [\%] $\uparrow$} \\
    \cmidrule(lr){2-4}
    \cmidrule(lr){5-7}
    & @5\degree & @10\degree & @20\degree & @5\degree & @10\degree & @20\degree\\
    \midrule
    Ours w/o multi-view & 24.9 & 42.5 & 59.6 & 60.7 & 75.3 & 85.0 \\
    Ours w/o end-to-end & 23.7 & 40.4 & 56.8 & 57.5 & 73.7 & 84.4 \\
    Ours & \textbf{26.9} & \textbf{45.6} & \textbf{63.0} & \textbf{64.2} & \textbf{78.8} & \textbf{87.7} \\
    \bottomrule
  \end{tabular}
  }
\end{center}
\caption{Ablation study on multi-view indoor pose estimation on ScanNet.}
\label{tab:ablation_scannet}
\begin{center}
\resizebox{\linewidth}{!}{
  \begin{tabular}{l >{\centering\arraybackslash}p{0.8cm} >{\centering\arraybackslash}p{0.8cm} >{\centering\arraybackslash}p{0.8cm} >{\centering\arraybackslash}p{0.8cm} >{\centering\arraybackslash}p{0.8cm} >{\centering\arraybackslash}p{0.8cm}}
    \toprule
    & \multicolumn{3}{c}{Transl.\ error AUC [\%] $\uparrow$} & \multicolumn{3}{c}{Rot.\ error AUC [\%] $\uparrow$} \\
    \cmidrule(lr){2-4}
    \cmidrule(lr){5-7}
    & @5\degree & @10\degree & @20\degree & @5\degree & @10\degree & @20\degree\\
    \midrule
    Ours w/o multi-view & 50.2 & 60.9 & 70.5 & 64.4 & 75.7 & 84.1 \\
    Ours w/o end-to-end & 49.9 & 60.8 & 70.5 & 61.6 & 74.7 & 84.2 \\
    Ours & \textbf{52.1} & \textbf{63.0} & \textbf{72.5} & \textbf{66.7} & \textbf{77.8} & \textbf{85.9} \\
    \bottomrule
  \end{tabular}
  }
  \end{center}
\caption{Ablation study on multi-view outdoor pose estimation on MegaDepth.}
\label{tab:ablation_megadepth}
\end{table}

\paragraphNoSpace{Variable Image Overlap.} \cref{tab:multi_view_pose_scannet_overlap} extends the multi-view pose estimation evaluation to a setting with reduced image overlap. It shows that our method achieves better pose estimation results than the baselines also in this setting. 
\begin{table}[htb]
\begin{center}
\resizebox{\linewidth}{!}{
\begin{tikzpicture}[very thick,squarednode/.style={rectangle, draw=none, fill=white, very thin, minimum size=2mm, text opacity=1,fill opacity=0}]
        \node[anchor=south west,inner sep=0] (image) at (0,0)
  {\begin{tabular}{@{}ll >{\centering\arraybackslash}p{0.8cm} >{\centering\arraybackslash}p{0.8cm} >{\centering\arraybackslash}p{0.8cm} >{\centering\arraybackslash}p{0.8cm} >{\centering\arraybackslash}p{0.8cm} >{\centering\arraybackslash}p{0.8cm}}
    \toprule
    && \multicolumn{3}{c}{Transl.\ error AUC [\%] $\uparrow$} & \multicolumn{3}{c}{Rot.\ error AUC [\%] $\uparrow$} \\
    \cmidrule(lr){3-5}
    \cmidrule(lr){6-8}
    && @5\degree & @10\degree & @20\degree & @5\degree & @10\degree & @20\degree\\
    \cmidrule(ll){2-8}
    &Mutual nearest neighbor & 8.5 & 17.8 & 31.0 & 33.0 & 48.4 & 62.8 \\
    &SuperGlue \cite{Sarlin2020SuperGlueLF} & 21.3 & 37.5 & 53.7 & 54.2 & 71.0 & 82.6 \\
    &LoFTR \cite{Sun2021LoFTRDL} & 20.6 & 36.9 & 53.7 & 57.3 & 72.0 & 82.0 \\
    &COTR \cite{Jiang2021COTRCT} cross-dataset & 10.9 & 22.4 & 36.9 & 38.8 & 53.6 & 66.3 \\
    &3DG-STFM \cite{Mao20223DGSTFM} & 22.0 & 38.7 & 55.5 & 57.0 & 72.7 & 83.0 \\
    &Ours & \textbf{26.9} & \textbf{45.6} & \textbf{63.0} & \textbf{64.2} & \textbf{78.8} & \textbf{87.7} \\
    \cmidrule(ll){2-8}
    &Mutual nearest neighbor & 3.4 & 8.1 & 16.9 & 12.7 & 23.6 & 38.1 \\
    &SuperGlue \cite{Sarlin2020SuperGlueLF} & 15.8 & 29.1 & 44.3 & 34.6 & 52.1 & 67.3 \\
    &LoFTR \cite{Sun2021LoFTRDL} & 15.8 & 28.5 & 43.1 & 35.6 & 51.6 & 65.1 \\
    &COTR \cite{Jiang2021COTRCT} cross-dataset & 5.4 & 11.9 & 22.2 & 17.4 & 29.0 & 42.6 \\
    &3DG-STFM \cite{Mao20223DGSTFM} & 15.4 & 28.1 & 43.0 & 34.3 & 50.3 & 64.5 \\
    &Ours & \textbf{20.9} & \textbf{36.6} & \textbf{53.0} & \textbf{42.8} & \textbf{60.0} & \textbf{73.6} \\
    \bottomrule
  \end{tabular}};
    \begin{scope}[x=(image.south east),y=(image.north west)]
        \node[squarednode,rotate=90,text=teal] at (0.005, 0.62) (a) {Overlap 1};
        \draw [pen colour={teal}, decorate, decoration = {calligraphic brace}] (0.03,0.42) --  (0.03,0.81);
        \node[squarednode,rotate=90,text=purple] at (0.005, 0.21) (a) {Overlap 2};
        \draw [pen colour={purple}, decorate, decoration = {calligraphic brace}] (0.03,0.02) --  (0.03,0.4);
        \end{scope}
    \end{tikzpicture}
  }
\end{center}
\caption{Multi-view indoor pose estimation using variable image overlap (range 1: \textcolor{teal}{$[0.4, 0.8]$}, range 2: \textcolor{purple}{$[0.25, 0.5]$}) on ScanNet; ``cross-dataset'' indicates that COTR was trained on MegaDepth.}
\label{tab:multi_view_pose_scannet_overlap}
\end{table}
\section{Qualitative Results}
\label{sec:qualitative_suppl}
\begin{figure*}[tb]
\begin{center}
    \begin{tikzpicture}[squarednode/.style={rectangle, draw=none, fill=white, very thin, minimum size=2mm, text opacity=1,fill opacity=0.}]
        \node[anchor=south west,inner sep=0] (image) at (0,0)
        {\includegraphics[width=\linewidth]{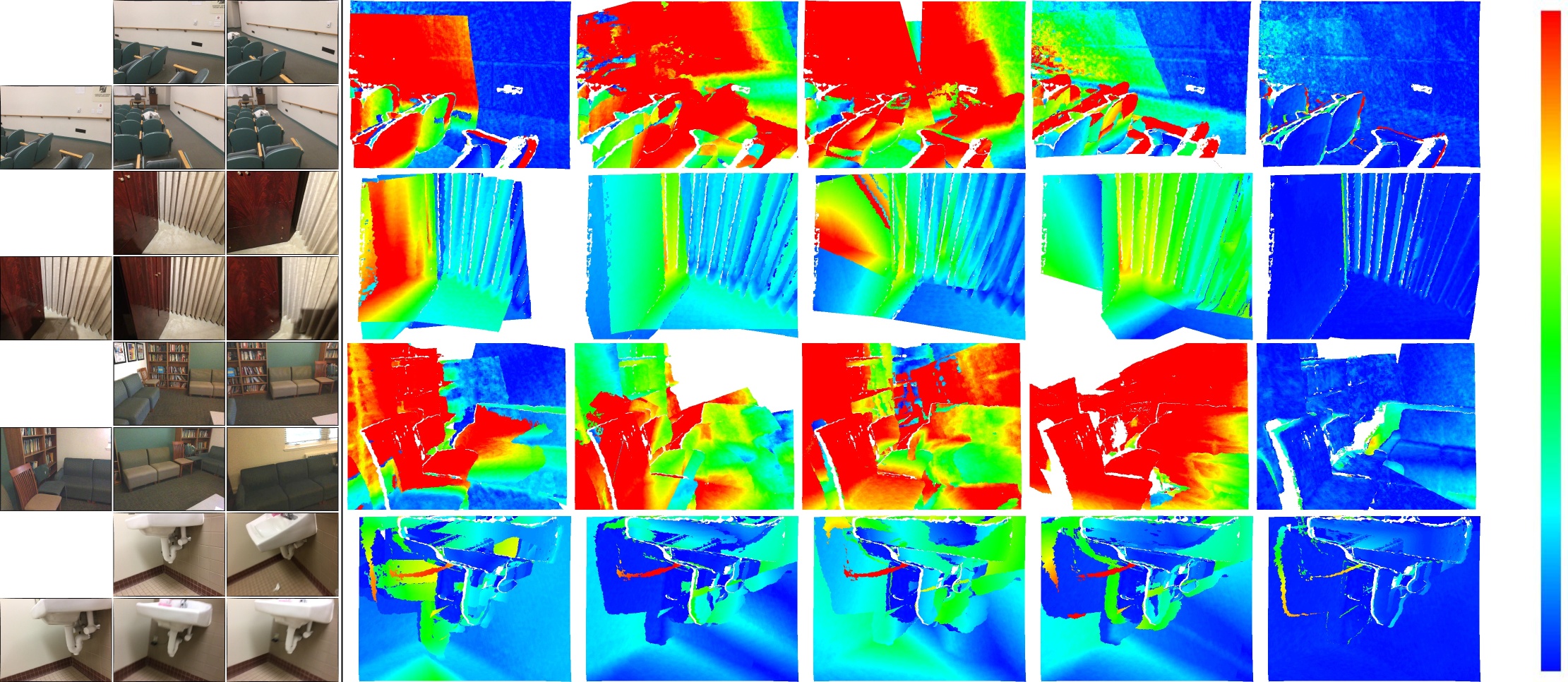}};
        \begin{scope}[x=(image.south east),y=(image.north west)]
        \node[squarednode] at (0.035, 0.94) (a) {\bf 1};
        \node[squarednode] at (0.035, 0.685) (b) {\bf 2};
        \node[squarednode] at (0.035, 0.435) (a) {\bf 3};
        \node[squarednode] at (0.035, 0.185) (b) {\bf 4};
        \node[squarednode] at (0.11, -0.04) (g) {Input 5-tuples};
        \node[squarednode] at (0.295, -0.04) (g) {SuperGlue \cite{Sarlin2020SuperGlueLF}};
        \node[squarednode] at (0.44, -0.04) (g) {LoFTR \cite{Sun2021LoFTRDL}};
        \node[squarednode] at (0.585, -0.04) (g) {COTR \cite{Jiang2021COTRCT}};
        \node[squarednode] at (0.73, -0.04) (g) {3DG-STFM \cite{Mao20223DGSTFM}};
        \node[squarednode] at (0.87, -0.04) (g) {Ours};
        \node[squarednode] at (0.965, 0.983) (g) {1m};
        \node[squarednode] at (0.965, 0.02) (g) {0m};
        \end{scope}
    \end{tikzpicture}
\end{center}
\caption{Reprojection error (right) for estimated camera poses on ScanNet 5-tuples (left). With multi-view matching and end-to-end training, our method successfully handles challenging pose estimation scenarios, while baselines have severe camera pose errors.}
\label{fig:scannet_results_suppl}
\end{figure*}

\begin{figure*}[tb]
\begin{center}
    \begin{tikzpicture}[squarednode/.style={rectangle, draw=none, fill=white, very thin, minimum size=2mm, text opacity=1,fill opacity=0.}]
        \node[anchor=south west,inner sep=0] (image) at (0,0)
        {\includegraphics[width=\linewidth]{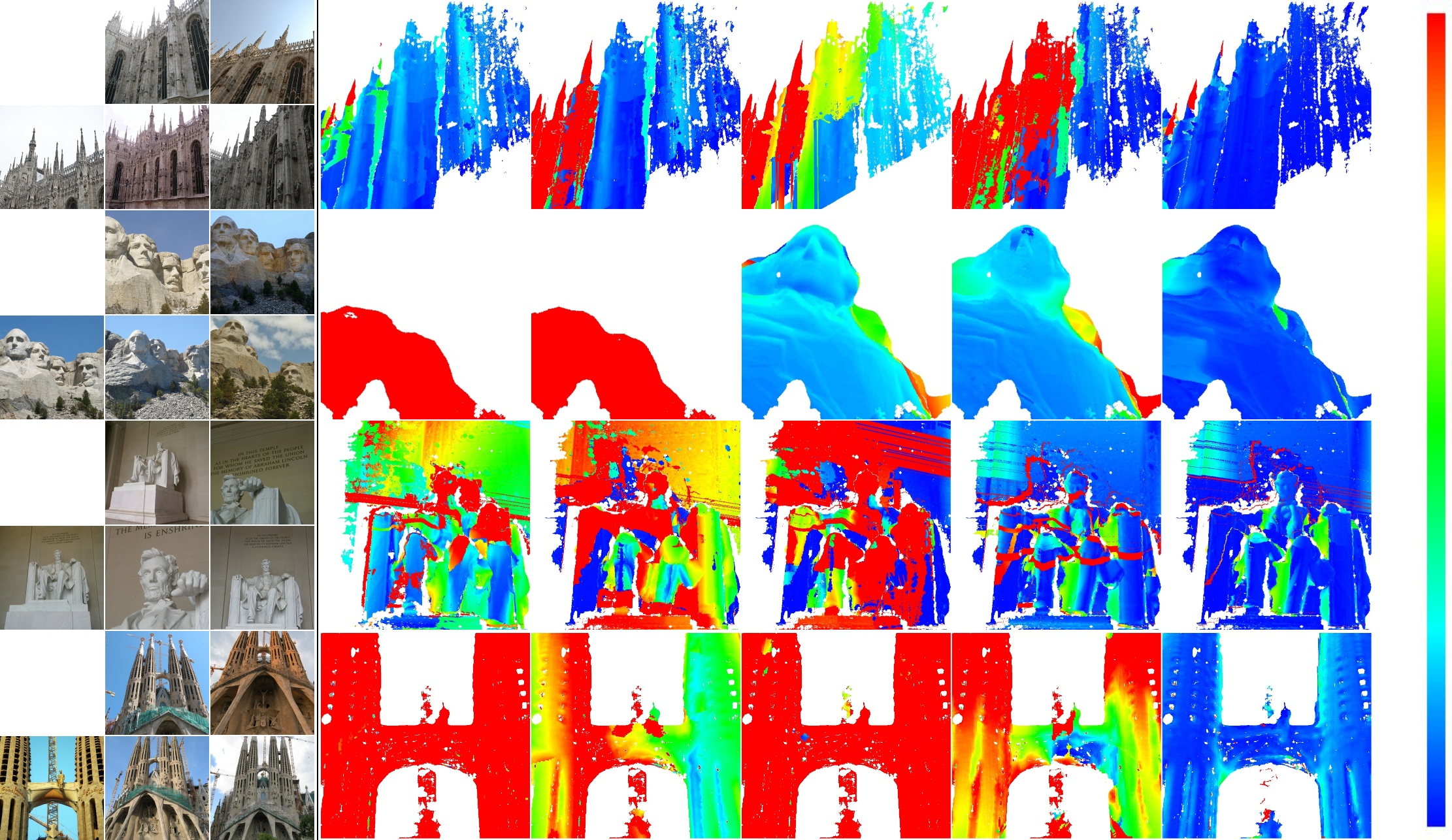}};
        \begin{scope}[x=(image.south east),y=(image.north west)]
        \node[squarednode] at (0.035, 0.94) (a) {\bf 1};
        \node[squarednode] at (0.035, 0.685) (b) {\bf 2};
        \node[squarednode] at (0.035, 0.435) (a) {\bf 3};
        \node[squarednode] at (0.035, 0.185) (b) {\bf 4};
        \node[squarednode] at (0.11, -0.03) (g) {Input 5-tuples};
        \node[squarednode] at (0.29, -0.03) (g) {SuperGlue \cite{Sarlin2020SuperGlueLF}};
        \node[squarednode] at (0.44, -0.03) (g) {LoFTR \cite{Sun2021LoFTRDL}};
        \node[squarednode] at (0.58, -0.03) (g) {COTR \cite{Jiang2021COTRCT}};
        \node[squarednode] at (0.725, -0.03) (g) {3DG-STFM \cite{Mao20223DGSTFM}};
        \node[squarednode] at (0.87, -0.03) (g) {Ours};
        \node[squarednode] at (0.965, 0.983) (g) {1};
        \node[squarednode] at (0.965, 0.02) (g) {0};
        \end{scope}
    \end{tikzpicture}
    \end{center}
\caption{Reprojection error (right) for estimated camera poses on MegaDepth 5-tuples (left). Through multi-view matching and end-to-end training, our method successfully estimates camera poses in challenging outdoor scenarios, while baselines show misalignment. Reprojection errors are visualized in the MegaDepth scaling.}
\label{fig:megadepth_results_suppl}
\end{figure*}

\begin{figure*}[tb]
\begin{center}
    \begin{tikzpicture}[squarednode/.style={rectangle, draw=none, fill=white, very thin, minimum size=2mm, text opacity=1,fill opacity=0.}]
        \node[anchor=south west,inner sep=0] (image) at (0,0)
        {\includegraphics[width=\linewidth]{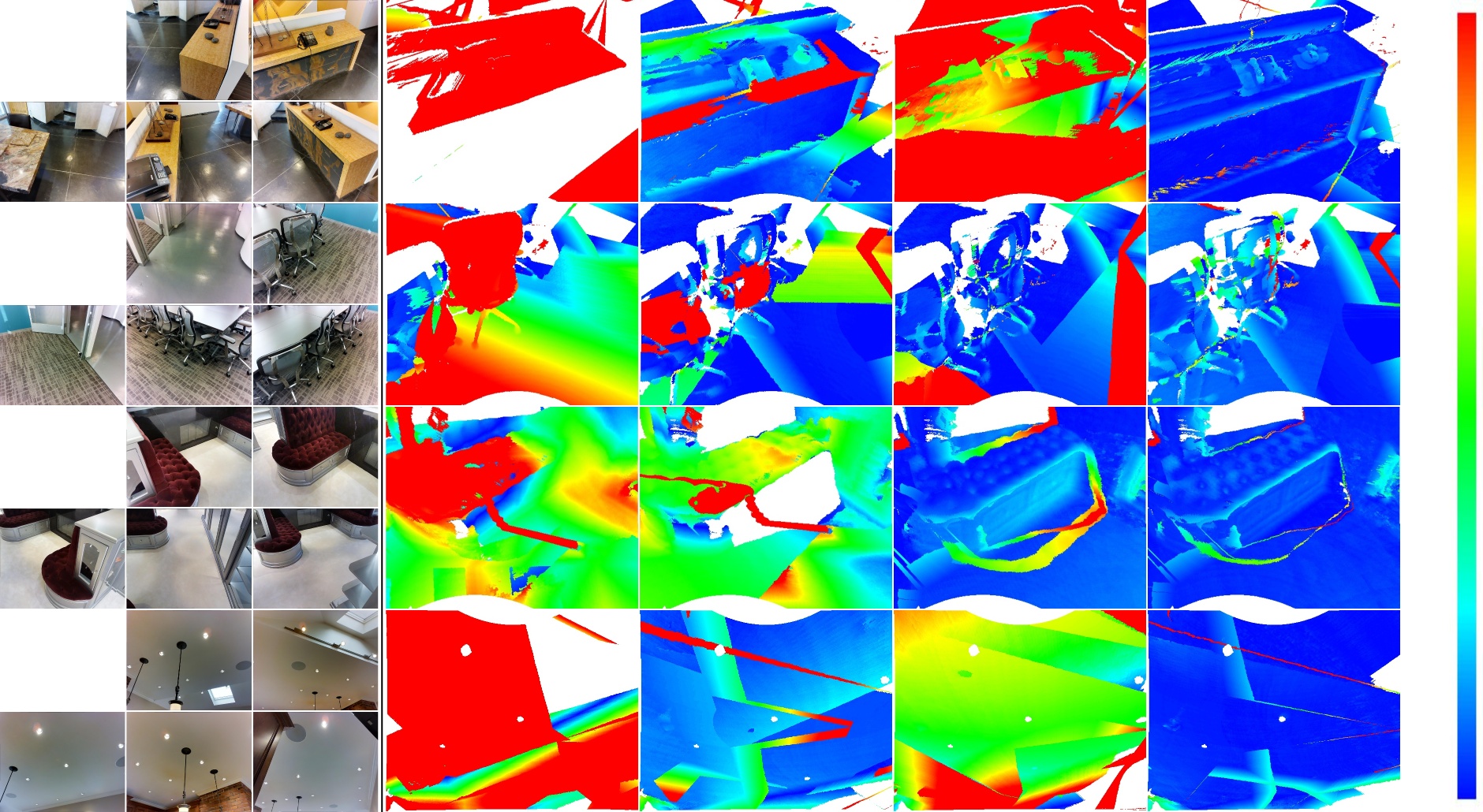}};
        \begin{scope}[x=(image.south east),y=(image.north west)]
        \node[squarednode] at (0.042, 0.944) (a) {\bf 1};
        \node[squarednode] at (0.042, 0.692) (b) {\bf 2};
        \node[squarednode] at (0.042, 0.441) (a) {\bf 3};
        \node[squarednode] at (0.042, 0.189) (b) {\bf 4};
        \node[squarednode] at (0.128, -0.03) (g) {Input 5-tuples};
        \node[squarednode] at (0.345, -0.03) (g) {SuperGlue \cite{Sarlin2020SuperGlueLF}};
        \node[squarednode] at (0.515, -0.03) (g) {Ours w/o multi-view};
        \node[squarednode] at (0.69, -0.03) (g) {Ours w/o end-to-end};
        \node[squarednode] at (0.86, -0.03) (g) {Ours};
        \node[squarednode] at (0.965, 0.983) (g) {1m};
        \node[squarednode] at (0.965, 0.02) (g) {0m};
        \end{scope}
    \end{tikzpicture}
\end{center}
\caption{Reprojection error (right) for estimated camera poses on Matterport3D 5-tuples (left). Our complete method improves camera alignment over the ablated versions and SuperGlue, showing the importance of multi-view matching and end-to-end training.}
\label{fig:matterport_results_suppl}
\end{figure*}

\cref{fig:scannet_results_suppl,fig:megadepth_results_suppl,fig:matterport_results_suppl} show additional qualitative results on ScanNet, MegaDepth and Matterport3D. Lower reprojection errors demonstrate that our matches give rise to more accurate pose estimation, even in texture-less areas (e.g., \cref{fig:scannet_results_suppl} sample 2) or across strong appearance changes (e.g., \cref{fig:megadepth_results_suppl} sample 1). 
\section{Cross-Dataset Results}
\label{sec:cross_dataset}
\begin{table}[htb]
\begin{center}
\resizebox{0.6\linewidth}{!}{
\begin{tabular}{lccc}
\toprule
& \multicolumn{3}{c}{Pose error AUC [\%] $\uparrow$} \\
\cmidrule(lr){2-4}
& @5\degree & @10\degree & @20\degree \\
\midrule
SuperGlue \cite{Sarlin2020SuperGlueLF} & 38.7 & 59.1 & 75.8 \\
LoFTR \cite{Sun2021LoFTRDL} & 43.5 & 63.5 & 78.6 \\
COTR \cite{Jiang2021COTRCT} & 34.4 & 54.7 & 71.8 \\
3DG-STFM \cite{Mao20223DGSTFM} & 43.4 & 63.4 & 78.4 \\
Ours & \textbf{46.7} & \textbf{65.4} & \textbf{79.3} \\
\bottomrule
\end{tabular}
}
\end{center}
\caption{Cross-dataset evaluation on two-view pose-estimation on YFCC100M. Models trained on MegaDepth.}
\label{tab:cross_dataset_yfcc100m}
\begin{center}
\resizebox{0.6\linewidth}{!}{
\begin{tabular}{lccc}
\toprule
& \multicolumn{3}{c}{Pose error AUC [\%] $\uparrow$} \\
\cmidrule(lr){2-4}
& @5\degree & @10\degree & @20\degree \\
\midrule
SuperGlue \cite{Sarlin2020SuperGlueLF} & 16.7 & 33.7 & 51.1 \\
LoFTR \cite{Sun2021LoFTRDL} & 17.7 & 34.7 & 51.1 \\
COTR \cite{Jiang2021COTRCT} & 11.8 & 26.5 & 42.5 \\
3DG-STFM \cite{Mao20223DGSTFM} & 16.1 & 32.3 & 49.2 \\
Ours & \textbf{18.8} & \textbf{36.4} & \textbf{52.8} \\
\bottomrule
\end{tabular}
}
\end{center}
\caption{Cross-dataset evaluation on two-view pose-estimation on ScanNet. Models trained on MegaDepth.}
\label{tab:cross_dataset_scannet}
\end{table}
\cref{tab:cross_dataset_yfcc100m,tab:cross_dataset_scannet} list cross-dataset results on two-view pose estimation, where the models are trained on MegaDepth and tested on YFCC100M~\cite{yfcc100m} and ScanNet. It shows that our method is able to transfer to different datasets. 
\section{Matching Metrics}
\label{sec:tradoff_p_auc}
Following the detector-based method SuperGlue, we compute precision (P) and matching score (MS) \cite{Sarlin2020SuperGlueLF}. 
Our end-to-end approach learns matching and outlier filtering in one step, hence, in contrast to the baselines, it does not need outlier filtering with RANSAC to estimate poses. \cref{tab:match_metrics_scannet} shows that we achieve comparable or higher precision and matching score than SuperGlue with RANSAC. 
\begin{table}[h]
\begin{center}
\resizebox{0.9\linewidth}{!}{
\begin{tabular}{llc@{\,\,\,\,}l@{\,\,\,\,}l@{}}
    \toprule
    & & RANSAC & P [\%] $\uparrow$ & MS [\%] $\uparrow$ \\
    \midrule
    SuperGlue \cite{Sarlin2020SuperGlueLF} & 2-view & \cmark & 93.8 (91.3) & 19.3 (38.6) \\
    Ours & 4-view & \xmark & \textbf{94.0} & 19.6 \\
    Ours & 5-view & \xmark & \textbf{94.0} & 19.4 \\
    Ours & 6-view & \xmark & 93.9 & \textbf{19.8} \\
    \bottomrule
  \end{tabular}
  }
\end{center}
\caption{Matching metrics on ScanNet. Our end-to-end method learns feature matching and outlier filtering in one step, hence, it does not require RANSAC and yields matches of similar or higher precision and matching score compared to SuperGlue with RANSAC. Parentheses indicate SuperGlue metrics w/o RANSAC.}
\label{tab:match_metrics_scannet}
\end{table}

\begin{figure}[htb]
\begin{center}
\begin{tikzpicture}[squarednode/.style={rectangle, draw=white, fill=white, very thin, minimum size=2mm, text opacity=1,fill opacity=0,draw opacity=0}]
        \node[anchor=south west,inner sep=0] (image) at (0,0)
        {
\includegraphics[width=0.65\linewidth,trim={0cm 0cm 0cm 0.2cm},clip]{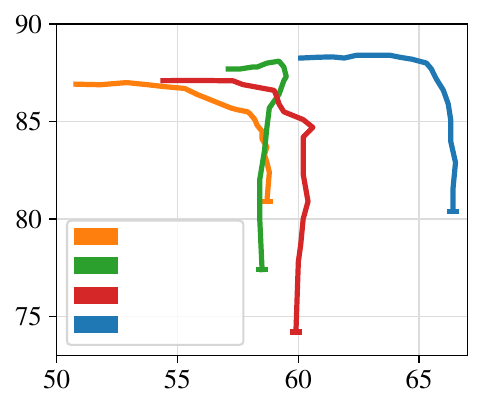}};
        \begin{scope}[x=(image.south east),y=(image.north west)]
        \small
        \node[squarednode]  at (0.38, 0.43) (a) {\scriptsize SuperGlue};
        \node[squarednode] at (0.38, 0.359) (a) {\scriptsize LoFTR};
        \node[squarednode] at (0.38, 0.287) (a) {\scriptsize 3DG-STFM};
        \node[squarednode] at (0.38, 0.212) (a) {\scriptsize Ours};
        \node[squarednode,rotate=90] at (-0.04, 0.55) (b) {Precision [\%]};
        \node[squarednode] at (0.55, -0.03) (b) {Pose error AUC@20\degree [\%]};
        \node[squarednode,rotate=-2] at (0.65, 0.245) (b) {\LARGE $\uparrow$};
        \node[squarednode] at (0.815, 0.26) (b) {\scriptsize increase conf.};
        \node[squarednode] at (0.815, 0.21) (b) {\scriptsize thresh. from 0};
        \end{scope}
\end{tikzpicture}
\end{center}
\caption{Trade-off between matching precision and pose estimation performance for variable confidence thresholds on ScanNet. Our matching results are both, of high precision and of high value for pose estimation.}
\label{fig:precision}
\end{figure}
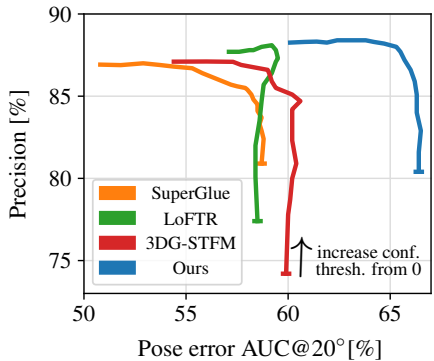

This evaluation (\cref{tab:match_metrics_scannet}) is not defined for the detector-free methods (as explained in \cite{Sun2021LoFTRDL}), therefore, we provide an alternative evaluation, which is applicable to the detector-free methods: \cref{fig:precision} visualizes the trade-off between the precision of matches and the pose estimation performance for increasing confidence thresholds (lower bound) starting at 0 until precision saturates. 
The curves are computed on the ScanNet image pairs from two-view pose estimation (main paper Section 4.1). 
Clearly, our method produces matching configurations with the best trade-off between precision and value for pose estimation. The baseline COTR does not provide confidences, hence its curve boils down to a point: 76.8\% precision at AUC@20\degree of 42.5\%. 
\section{Matching Runtime}
\label{sec:matching_runtime}
\cref{tab:runtime_scannet} lists the matching runtime 
for increasing number of views, measured on a Nvidia GeForce RTX 2080. It shows that joint multi-view matching is faster than matching the corresponding pairs with SuperGlue. 
The savings stem from fewer intra-frame, self-attention
GNN messages in multi-view matching compared to pairwise (see \cref{sec:gnn_messages}). 
\begin{table}[htb]
\begin{center}
\resizebox{\linewidth}{!}{
  \begin{tabular}{lccccc}
    \toprule
    & 2-view & 4-view & 5-view & 6-view & 8-view \\
    & $\widehat{=}$ 1 pair & $\widehat{=}$ 6 pairs & $\widehat{=}$ 10 pairs & $\widehat{=}$ 15 pairs & $\widehat{=}$ 28 pairs \\
    \midrule
    SuperGlue \cite{Sarlin2020SuperGlueLF} & \textbf{45}ms & 190ms & 315ms & 470ms & 849ms\\
    Ours & \textbf{45}ms & \textbf{181}ms & \textbf{260}ms & \textbf{352}ms & \textbf{589}ms \\
    \bottomrule
  \end{tabular}
  }
\end{center}
\caption{Matching runtime (excluding SuperPoint) for variable number of views on ScanNet.}
\label{tab:runtime_scannet}
\end{table}
\section{Cross-Attention Visualization}
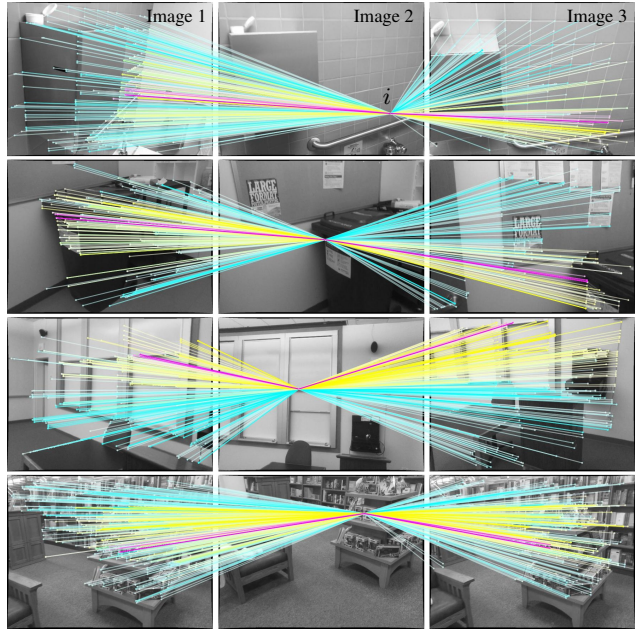
\begin{figure}[tb]
\begin{center}
\begin{tikzpicture}[squarednode/.style={rectangle, draw=white, fill=white, very thin, minimum size=2mm, text opacity=1,fill opacity=0,draw opacity=0}]
        \node[anchor=south west,inner sep=0] (image) at (0,0)
        {
\includegraphics[width=1.0\linewidth,trim={0cm 0cm 0cm 0cm},clip]{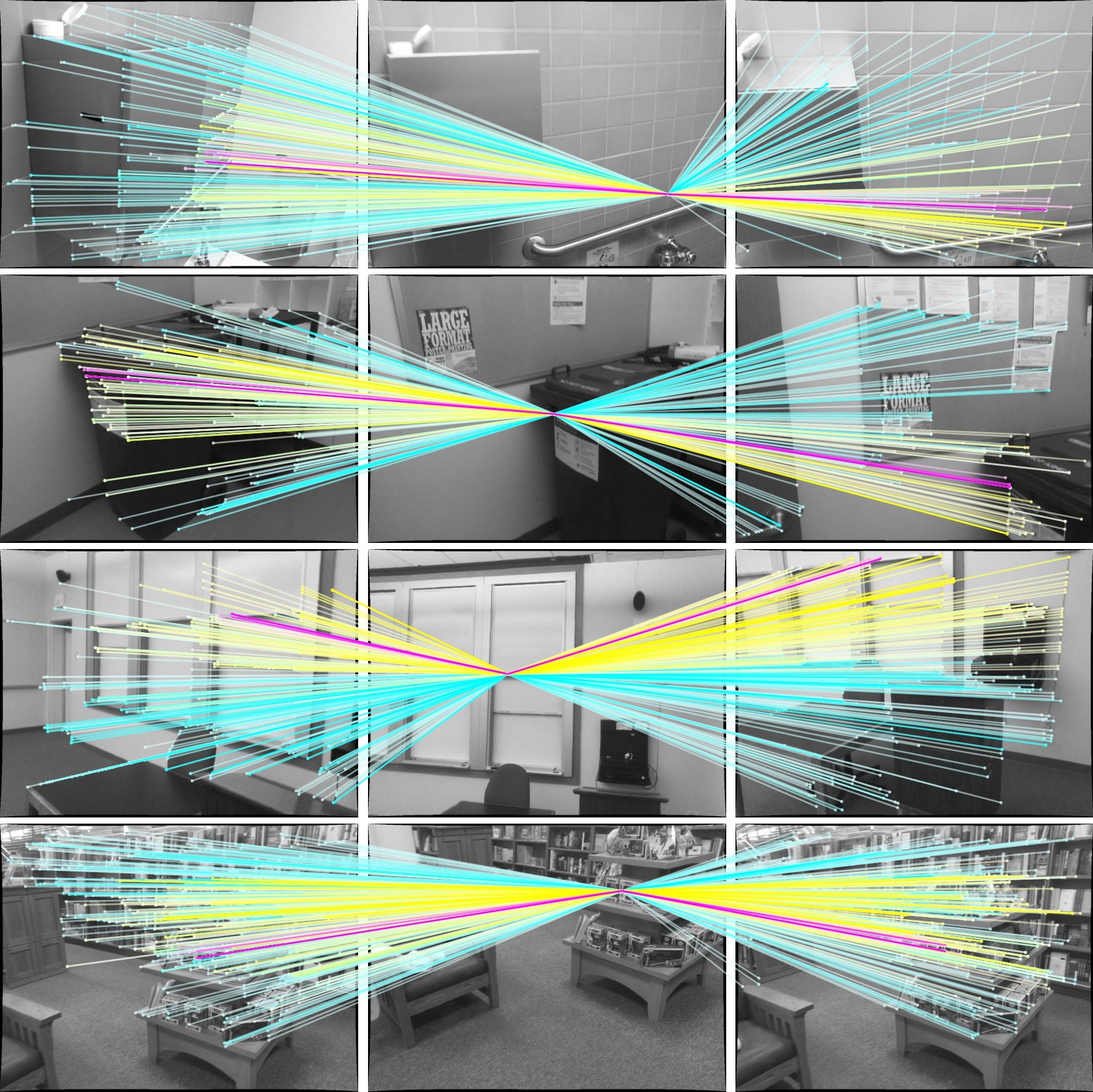}};
        \begin{scope}[x=(image.south east),y=(image.north west)]
        \small
        \node[squarednode] at (0.605, 0.85) (a) {\small \textcolor{black}{$i$}};
        \node[squarednode] at (0.27, 0.975) (a) {\scriptsize Image 1};
        \node[squarednode] at (0.6, 0.975) (a) {\scriptsize Image 2};
        \node[squarednode] at (0.94, 0.975) (a) {\scriptsize Image 3};
        \end{scope}
\end{tikzpicture}
\end{center}
\caption{\textcolor{cyan}{Early}/{\setlength{\fboxsep}{1.pt}\colorbox{yellow}{mid}}/\textcolor{magenta}{late} layer cross-attention weights as opacity. Keypoint $i$ in image 2 first interacts with spread points in images 1 and 3, then focuses around the match in middle and late cross-attention layers.}
\label{fig:cross_attention}
\end{figure}

\cref{fig:cross_attention} visualizes cross-attention weights. In early layers keypoints interact with spread keypoints in the other images. In later layers, cross-attention more and more focuses on the region of the matching keypoint. 
\section{Training with Bundle Adjustment}
\label{sec:ba_training_suppl}
We found that adding bundle adjustment in the end-to-end training, compared to training with 
weighted eight-point alone, leads to a minor improvement in the pose error AUC (\cref{tab:e2e_training_with_ba})---hence, we favored the simpler training procedure with weighted eight-point alone. At test time, however, the pose refinement with bundle adjustment is highly beneficial as shown in the experiment section of the main paper. 
\begin{table}[htb]
\begin{center}
\resizebox{\linewidth}{!}{
\begin{tabular}{@{}lccccc@{}}
\toprule
& \multirow{2}{0.27\linewidth}{\centering weight. 8-point training} & \multirow{2}{0.27\linewidth}{\centering bundle adjust. training} & \multicolumn{3}{c}{Pose error AUC [\%] $\uparrow$} \\
\cmidrule(lr){4-6}
& & & @5\degree & @10\degree & @20\degree \\
\midrule
Ours & \cmark & \xmark & 25.7 & 47.2 & 66.4 \\
Ours & \cmark & \cmark & \textbf{26.0} & \textbf{47.6} & \textbf{66.7} \\
\bottomrule
\end{tabular}
}
\end{center}
\caption{End-to-end training with weighted 8-point and bundle adjustment on ScanNet.}
\label{tab:e2e_training_with_ba}
\end{table}
\section{Number of GNN Messages}
\label{sec:gnn_messages}
\cref{tab:gnn_messages} shows that jointly matching $N$ images in a single graph reduces the number of GNN messages along self-edges compared to separately matching the corresponding $P=\sum_{n=1}^{N-1}n$ pairs. 
E.g., consider matching 5 images with $K$ keypoints each, either (A) jointly in a single match graph or (B) matching the 10 possible pairs. 
In each layer, (A) computes self-attention for 5 images, hence $5K^2$ GNN messages (B) computes self-attention for 10 pairs, i.e., 20 images, hence  $20K^2$  GNN messages. 
The number of messages along cross-edges is the same in pairwise and joint matching. 
\begin{table}[htb]
\begin{center}
\resizebox{\linewidth}{!}{
\begin{tabular}{lcc}
\toprule
& \multicolumn{2}{c}{Number of GNN messages} \\
\cmidrule(lr){2-3}
& along self-edges & along cross-edges \\
\midrule
Pairwise matching & $2PK^2$ & $N(N-1)K^2$ \\
Joint matching & $NK^2$ & $N(N-1)K^2$ \\
\bottomrule
\end{tabular}
}
\end{center}
\caption{Number of GNN messages per layer for matching $N$ images, each  with $K$ keypoints, as $P$ individual image pairs versus joint matching in a single graph.}
\label{tab:gnn_messages}
\end{table}
\section{Architecture Details}
\label{sec:architecture_suppl}
Our multi-view matching network is inspired by the SuperGlue \cite{Sarlin2020SuperGlueLF} architecture. 

\paragraphNoSpace{Keypoint Encoder.}
The input visual descriptors from SuperPoint \cite{DeTone2018SuperPointSI} have size $D = 256$. The graph nodes equally have an embedding size of $D$. Hence, the keypoint encoder $F_{\mathrm{encode}}$ maps a keypoint's image coordinates and confidence score to $D$ dimensions. It is a MLP,  composed of five layers with 32, 64, 128, 256 and $D$ channels. Each layer, except the last, uses batch normalization and ReLU activation. 

\paragraphNoSpace{Graph Attention Network.} 
We found that multi-view matching benefits from more information flow along cross-edges compared to self-edges. Hence, the GNN has 7 self-attention layers, each followed by three cross-attention layers. In the two-view setting and on MegaDepth---due to limited amount of data---we use a smaller network size with 9 self- and 9 cross-attention layers in alternating fashion. 
The attentional aggregation of incoming messages from other nodes uses multi-head attention with four heads. The resulting messages have size $D$, like the node embeddings. 
The MLP $F_{\mathrm{update}}$, which computes the update to the receiving node, operates on the concatenation of the current node embedding with the incoming message. It has two layers with $2D$ and $D$ channels. Batch normalization and ReLU activation are employed between the two layers. 

\paragraphNoSpace{Partial Assignment.}
We use 100 iterations of the Sinkhorn algorithm to determine the partial assignment matrices. 

\paragraphNoSpace{Confidence MLP.} 
$F_\mathrm{conf\_3}$ merges the final node descriptors of matching keypoints---i.e., it operates on the concatenated match descriptors and applies two linear layers with $2D$ and $D$ channels. $F_\mathrm{conf\_2}$ lifts the corresponding partial assignment score to descriptor space through two linear layers with $D$ channels each. 
The $D$-dimensional output embeddings of $F_\mathrm{conf\_2}$ and $F_\mathrm{conf\_3}$ are summed and fed into $F_\mathrm{conf\_1}$, which is a final linear layer with sigmoid activation that reduces to a single channel, the matching confidence. 
All layers in $F_\mathrm{conf\_2}$ and  $F_\mathrm{conf\_3}$ use batch normalization and ReLU activation. 

\paragraphNoSpace{Pose Optimization.}
The camera poses are optimized by conducting \mbox{$T=5$} Gauss-Newton updates at training time and \mbox{$T=10$} at test time. The damping factor $\beta$ is initially set to 0.1. It is divided by a factor of 3.5 if the magnitude of the residual vector decreases, conversely, it is multiplied by a factor of 1.5 if the magnitude of the residual vector increases. 
\section{Training Details}
\label{sec:training_suppl}
\paragraphNoSpace{Two-Stage Training.} 
Our end-to-end pipeline is trained in two stages. The first stage uses the loss term on the matching result $\mathcal{L}_{\mathrm{match}}$. The second stage additionally applies the pose loss $\mathcal{L}_{\mathrm{pose}}$. Stage 1 is trained until the validation match loss converges, stage 2 until the validation pose loss converges. On ScanNet/ Matterport3D/ MegaDepth the training takes 32/ 343/ 143 epochs for stage 1 and 40/ 365/ 126 epochs for stage 2. We found that the training on Matterport3D and MegaDepth benefits from initializing the network weights to the weights after the first training stage on ScanNet, where most data is available. During stage 2 we linearly increase the weight of $\mathcal{L}_{\mathrm{pose}}$ from 0 to 242/ 585/ 345 on ScanNet/ Matterport3D/ MegaDepth, while linearly decreasing the weight of $\mathcal{L}_{\mathrm{match}}$ from 1 to 0.01, over a course of 40000 iterations. The balancing factor of the rotation term $\lambda_{\mathrm{rot}}$ is set to 3.0/ 1.2/ 2.0 on ScanNet/ Matterport3D/ MegaDepth. 
We use the Adam optimizer \cite{Kingma2015AdamAM} with learning rate 0.0001. 
The learning rate is exponentially decayed with a factor of 0.999992 starting after 100k iterations. 

\paragraphNoSpace{Ground Truth Generation.} 
The ground truth matches $\mathcal{T}_{ab}$ and sets of unmatched keypoints $\mathcal{U}_{ab}$, $\mathcal{V}_{ab}$ of an image pair are computed by projecting the detected keypoints from each image to the other, resulting in a reprojection error matrix. Keypoint pairs where the reprojection error is both minimal and smaller than 5 pixels in both directions are considered matches. Unmatched keypoints must have a minimum reprojection error greater than 15 pixels on the indoor datasets and greater than 10 pixels on MegaDepth. 

\paragraphNoSpace{Input Data.}
We train the multi-view model on 5-tuples, which are sampled based on overlap ranges. On ScanNet and Matterport3D, overlap is computed using the ground truth poses, depth maps and intrinsic parameters. Following prior work \cite{Sarlin2020SuperGlueLF,Sun2021LoFTRDL,Mao20223DGSTFM}, an overlap range of $[0.4, 0.8]$ is used on ScanNet. On Matterport3D, where view capture is much more sparse, we relax the overlap criterion to $[0.25, 0.8]$. 
On MegaDepth, the overlap between images is the portion of co-visible 3D points of the sparse reconstruction \cite{Sarlin2020SuperGlueLF,Dusmanu2019D2NetAT}, thus the overlap definition is different from the indoor datasets and not comparable. Overlap ranges $[0.1, 0.7]$ and $[0.1, 0.4]$ are used at train and test time, respectively \cite{Sarlin2020SuperGlueLF}. 
The network is trained with a batch size of 24 on indoor data and with a batch size of 4 on outdoor data. 
The image size is 480$\times$640 on ScanNet, 512$\times$640 on Matterport3D and 640$\times$640 on MegaDepth. 
The SuperPoint network is configured to detect keypoints with a non-maximum suppression radius of 4/ 3 on indoor/ outdoor data. 
On the indoor datasets we use 400 keypoints per image during training time: first, keypoints above a confidence threshold of 0.001 are sampled, second, if there are fewer than 400, the remainder is filled with random image points and confidence 0 as a data augmentation. On MegaDepth the same procedure is applied to sample 1024 keypoints using a confidence threshold of 0.005. At test time on indoor/ outdoor data, we use up to 1024/ 2048 keypoints above the mentioned confidence thresholds. 

\paragraphNoSpace{Dataset Split.}
On ScanNet and Matterport3D, we use the official dataset split. 
On MegaDepth, we follow the data split of prior work \cite{Sun2021LoFTRDL,Tyszkiewicz2020DISKLL,Mao20223DGSTFM} using scenes 0015 and 0022 for validation, scenes 0008, 0019, 0021, 0024, 0025, 0032, 0063 and 1589 for testing and the remaining scenes for training. Scenes with low quality depth maps are filtered out \cite{Tyszkiewicz2020DISKLL,Sun2021LoFTRDL,Jiang2021COTRCT,Mao20223DGSTFM}.
This way, on ScanNet/ Matterport3D/ MegaDepth we have 240k/ 20k/ 15k 5-tuples for training, 62k/ 2200/ 200 for validation and 1500/ 1500/ 1500 for testing. 
\section{Baseline Comparison Details}
\label{sec:baseline_suppl}
In the baseline comparison, we use the network weights provided by the authors of SuperGlue~\cite{Sarlin2020SuperGlueLF}, LoFTR~\cite{Sun2021LoFTRDL}, COTR~\cite{Jiang2021COTRCT} and 3DG-STFM~\cite{Mao20223DGSTFM}. 
There are SuperGlue, LoFTR and 3DG-STFM models trained on ScanNet and on MegaDepth, as well as a COTR model trained on MegaDepth. We additionally train a SuperGlue model on Matterport3D and a SuperGlue model on MegaDepth using the above described dataset split, which is necessary as the provided model was trained on a train set that contains our test set, as well as the Image Matching Challenge scenes. 
For the baselines, SuperGlue, LoFTR, and 3DG-STFM, we use their default confidence thresholds---0.2 for all three---and verify that they benefit from this threshold. 
We found that our method predicts accurate confidences that do not require thresholding for weighted pose estimation. When using RANSAC for two-view pose estimation, we filter matches from our model w/o multi-view using a threshold of 0.02. 

In the multi-view evaluation we found that all methods benefit from a confidence-weighted bundle adjustment formulation on the inlier matches using Ceres solver (step (iv) in Section 4.2). 
Following \cite{lindenberger2021pixsfm}, we conduct the Image Matching Challenge (IMC)~\cite{jin2021image} multi-view evaluation on the scenes Reichstag, Sacre Coeur and St.\ Peter's Square. The above described MegaDepth dataset split ensures that these scenes do not overlap with the training set. 
Since the IMC protocol does not consider matches in a confidence-weighted manner, we apply a threshold of 0.06 on matches from our multi-view model. 

Following \cite{Sarlin2020SuperGlueLF}, matches are considered correct if the symmetric epipolar distance is smaller than $5\cdot 10^{-4}$ or $1\cdot 10^{-4}$ in the indoor and outdoor setting, respectively. 
\end{document}